\documentclass[pdflatex,sn-basic]{sn-jnl}
\usepackage[utf8]{inputenc}
\usepackage{graphicx}
\usepackage{amsmath,amssymb,amsfonts,amsthm,mathtools,mathrsfs,bm}
\usepackage{booktabs,multirow,makecell,adjustbox,tabularx}
\usepackage{placeins,rotating,float}
\usepackage[table,x11names]{xcolor}
\usepackage{url,enumitem,comment}
\usepackage[english]{babel}
\definecolor{RoyalBlue}{rgb}{0.25,0.41,0.88}
\definecolor{light-gray}{gray}{0.97}
\definecolor{qblack}{gray}{0.30}
\hypersetup{pdftitle={Risk-optimal active learning with informative abstention},pdfauthor={Fariborz Setoudehtazangi and Geoffrey J. McLachlan},bookmarksdepth=2,hypertexnames=false,colorlinks=true,linkcolor=blue,urlcolor=blue,citecolor=blue}
\setcitestyle{authoryear,round,aysep={},yysep={,},notesep={; }}
\allowdisplaybreaks[1]
\theoremstyle{thmstyleone}   

\newtheorem{theorem}{Theorem}[section]
\newtheorem{proposition}[theorem]{Proposition}

\theoremstyle{thmstylethree}  

\theoremstyle{thmstyletwo}  

\numberwithin{equation}{section}

\begin{document}
\title[Risk-optimal active learning with informative abstention]{Risk-optimal active learning with informative abstention}
\author[1]{\fnm{Fariborz} \sur{Setoudehtazangi}}\email{fariborz.setoudehtazangi@studenti.unipd.it}
\author*[2]{\fnm{Geoffrey J.} \sur{McLachlan}}\email{g.mclachlan@uq.edu.au}
\affil[1]{\orgdiv{Dipartimento di Scienze Statistiche},\ 
\orgname{Universit\`a di Padova},\ \orgaddress{\city{Padova},\ \country{Italy}}}
\affil[2]{\orgdiv{School of Mathematics and Physics},\ 
\orgname{The University of Queensland},\ \orgaddress{\city{Brisbane},\ \country{Australia}}}
\abstract{We study label acquisition when an oracle may abstain and its response
probability shares parameters with the classifier. A query can then
provide information through both a returned label and the response
outcome. Under a correctly specified parametric model, we derive the
classifier information after adjustment for response-model nuisance
parameters. Combining this information with the local curvature of
multiclass zero--one risk gives an acquisition criterion for minimizing
leading asymptotic excess risk. We characterize the optimal allocation
under a fixed query budget and show how query allocation affects
response identifiability. The oracle criterion reduces to
response-discounted acquisition when response probability does not
depend on the classifier parameters. Simulations examine the resulting
query allocations and isolate the contribution of the response-outcome
term through an ablation. A Consumer Complaints replay retains observed
human abstentions and reveals benchmark labels upon response.
Response-aware acquisition has lower mean test error than
response-discounted acquisition at the two largest budgets, with paired
differences significant after Holm correction. Entropy sampling performs
better at the smallest budget. The gains depend on the response
mechanism and query budget.}
\keywords{Active learning, Informative oracle response, Label acquisition,
Abstention, Fisher information, Classification risk}
\maketitle

\section*{Acknowledgements}

We are grateful to Cuong V. Nguyen of the Department of
Mathematical Sciences at Durham University, UK, for
generously taking the time to read the manuscript and
for his thoughtful comments. His observations on
adaptive sampling helped us clarify the assumptions
and proof of the two-stage acquisition result.

\section{Introduction}
\label{sec:introduction}

Obtaining a class label may require a human annotator, expert
assessment, diagnostic procedure, or user interaction. A query does
not always produce a label: an annotator may decline an ambiguous
case, an expert may find the available information insufficient,
or a user may not respond. This raises two questions for active
learning: which observations should be queried, and how should the
possibility and the observed pattern of response affect query
selection?

Classical active learning allocates a limited labeling budget by
selectively requesting labels
\citep{lewis1994sequential,lewis1994heterogeneous,settles2009active}.
Uncertainty sampling prioritizes observations with ambiguous class
probabilities, whereas information-based methods assess the value
of a label for parameter estimation or prediction
\citep{hoi2006batch,sourati2017asymptotic,chen2023firal}.
Other approaches target prediction error, predictive information,
or downstream decisions
\citep{roy2001toward,bickfordsmith2023prediction,
filstroff2024targeted,huang2026loss}.

Statistical approaches to sampling provide related foundations.
\citet{wang2018optimal} and \citet{yao2019optimal} derive optimal
subsampling procedures for logistic and softmax regression,
respectively. \citet{imberg2020optimal} study sampling schemes for
prediction-error reduction in unbiased active learning, while
\citet{bach2006active} develops active learning for potentially
misspecified generalized linear models. These methods address
different inferential and predictive objectives. The problem
considered here additionally involves a stochastic oracle response
whose distribution shares parameters with the classifier.

Imperfect labeling oracles have received substantial attention.
\citet{yan2016imperfect} study active learning when a labeler may
return an incorrect label or abstain, obtaining query-complexity
guarantees under assumptions on the noise and abstention processes.
\citet{fang2014uncertain} consider uncertain labeling knowledge,
allowing an oracle to indicate that it does not know a label.
\citet{amin2021abstentions} study a different form of abstention:
examples routed to a human labeler no longer require a prediction
from the machine. Their algorithm jointly learns a model and decides
when to request labels. In a Bayesian pool-based setting,
\citet{nguyen2022bayesian} develop acquisition rules that learn both
the classifier and an unknown abstention mechanism. Thus, neither
allowing an oracle to abstain nor learning its response mechanism
alongside the classifier is itself new.

A closely related approach is the biased-nonresponse framework of
\citet{robinson2024nonresponse}. Their expected-utility rule weights
an ordinary acquisition utility by the estimated probability of
response; an upper-confidence-bound extension accounts for
uncertainty in that estimate. This addresses the reduced value of
queries that are unlikely to return labels. It does not, however,
derive classifier information from the realized response outcome
through a shared-parameter likelihood. Likewise,
\citet{nguyen2022bayesian} use separate classification and abstention
models, rather than deriving the nuisance-adjusted Fisher
information of a classifier-linked response mechanism and weighting
it by local zero--one classification risk.

Likelihood-based work on informative label missingness motivates
this distinction. Under Rubin's framework, conditional independence
of missingness and the unobserved class given the features does not
by itself justify ignoring a mechanism that shares parameters with
the classification model \citep{rubin1976inference}.
\citet{ahfock2020paradox} show that uncertainty-dependent label
missingness can carry information about a discriminant rule.
More recent work develops information decompositions and connects
the resulting information to classification performance
\citep{wang2026learning,SetoudehtazangiMcLachlan2026a}.
These results concern the information available under a given
missingness mechanism. Here the investigator also controls which
observations are submitted to the oracle.

Our starting point is the classification-risk acquisition framework
of \citet{SetoudehtazangiMcLachlan2026b}, in which every requested
label is returned. Under its regularity conditions, leading
asymptotic excess zero--one risk depends on Fisher information
weighted by the local curvature \(H_R\) of the active Bayes decision
boundary. A label is therefore valuable according to whether its
information improves estimation in directions that affect
classification, rather than according to uncertainty or parameter
precision alone. We consider how this criterion changes when the
oracle may abstain.

Let \(Y\) denote the observed features, \(Z\) the class label,
\(A\) the investigator's query indicator, and \(R\) the oracle
response indicator. Conditional on a query, we model
\[
\Pr(R=1\mid A=1,Y=y)=r(y;\theta,\eta),
\]
where \(\theta\) governs the classification model and \(\eta\)
contains response-specific nuisance parameters. We assume that a
returned label is correct and that
\[
R\perp Z\mid(Y,A=1).
\]
The acquisition rule is known and parameter-free. Conditional on
the observed features, its realized query decision therefore carries
no likelihood information about the model parameters. The subsequent
response outcome can carry such information because its probability
depends on \(\theta\).

For the joint parameter
\(\phi=(\theta^\top,\eta^\top)^\top\), a query at \(y\) contributes
\[
K(y)
=
r(y)\widetilde J(y)
+
r(y)\{1-r(y)\}q(y)q(y)^\top,
\]
where \(\widetilde J(y)\) is the conditional class-label information
augmented by zero nuisance blocks, and
\[
q(y)=\nabla_\phi\operatorname{logit}r(y;\theta,\eta).
\]
The first term is information from a potentially returned label.
The second is information from observing whether the oracle responds.
When the response-model nuisance information is nonsingular,
eliminating \(\eta\) leaves a positive semidefinite response
contribution to efficient classifier information. This contribution
vanishes precisely when the classifier-related response score is
explained by the nuisance response score under the design-weighted
measure. Sharing parameters is therefore insufficient by itself:
useful response information must survive nuisance adjustment and
affect parameter directions relevant to classification.

Combining the query information with local risk curvature gives
the marginal acquisition value
\[
\psi_a(y)=\psi_a^L(y)+\psi_a^R(y),
\]
where \(a\) denotes the acquisition design.
The returned-label term \(\psi_a^L(y)\) accounts for the probability
of receiving an informative label. The response term
\(\psi_a^R(y)\) accounts for the classification value of the
response outcome itself. A query can consequently be informative
even when no class label is returned. When response probability
does not depend on the classifier parameters, the oracle criterion
reduces to response-discounted acquisition.

The joint information matrix is affine in the acquisition design.
The resulting optimization therefore retains the convex
trace-inverse structure of a linear optimal-design criterion
\citep{pukelsheim2006optimal} and admits a threshold characterization
under a fixed query budget. Query allocation must also preserve
sufficient variation for nonsingular response-model information.

We derive this rank condition and give a two-stage oracle-value
result for a pilot-selected design applied to fresh observations.
The risk conclusion requires uniform criterion consistency,
uniform asymptotic linearity in mean square, and a higher-moment
bound. The proof conditions on the pilot and then averages over it.
These conditions are not established for the sequential finite-pool
implementation used in the experiments.

The simulations examine oracle and adaptive acquisition under
noninformative and informative response mechanisms. An ablation
retains the full information matrix and joint estimation while
omitting the explicit response-outcome term from query scores.
It separates the effect of that term from changes in the
information matrix used by response-discounted acquisition.
The observed gains are small and depend on the response mechanism
and query budget.

We also use the Consumer Complaints annotations collected by
\citet{totakura2026crowd}. Each worker is treated as a separate
oracle. The replay preserves observed human response and abstention
patterns but reveals the benchmark class whenever a worker responds.
It therefore evaluates acquisition under human abstention without
modeling errors in returned labels. This comparison evaluates the
complete acquisition rules; it does not isolate the explicit
response-outcome term.

The contribution is a classification-risk acquisition criterion for
a shared-parameter query--response model. It values response
outcomes through their nuisance-adjusted information and their
effect on the classification boundary. The analysis identifies
conditions under which this additional information changes query
selection, while the experiments examine when those changes improve
classification.

Section~\ref{sec:model} introduces the query--response model and
derives its Fisher information.
Section~\ref{sec:efficient-information} characterizes the efficient
classifier information after nuisance adjustment.
Section~\ref{sec:risk} derives the classification-risk criterion and
its marginal query value. Subsequent sections examine optimal
allocation, response identifiability, adaptive acquisition, and a
binary logistic illustration. Simulations and the Consumer
Complaints replay follow, before the discussion of findings and
limitations.
\section{Statistical framework and informative oracle response}
\label{sec:model}

Let \(Y\in\mathcal Y\subseteq\mathbb R^p\) denote an observed feature vector and
\(Z\in\{1,\ldots,g\}\) its class label. We consider a regular parametric
classification model
\[
p_\theta(y,z)
=
p_\theta(y)\tau_z(y;\theta),
\qquad
\tau_k(y;\theta)
=
\Pr_\theta(Z=k\mid Y=y),
\]
where \(\theta\in\Theta\subseteq\mathbb R^d\) is the parameter governing the
classification model.

The feature vector is observed for every sampling unit. After observing
\(Y=y\), the investigator may request its class label. Let
\[
A=
\begin{cases}
1, & \text{if a label is requested},\\
0, & \text{otherwise},
\end{cases}
\]
and suppose
\begin{equation}
\Pr(A=1\mid Y=y,Z)=a(y),
\qquad 0\leq a(y)\leq1.
\label{eq:acquisition-design}
\end{equation}
The measurable function \(a:\mathcal Y\rightarrow[0,1]\) is the
\emph{acquisition design}. It is chosen by the investigator and is treated as
known and parameter-free.

A label request need not result in a returned label. Conditional on \(A=1\),
let
\[
R=
\begin{cases}
1, & \text{if the oracle returns a label},\\
0, & \text{if the oracle abstains}.
\end{cases}
\]
We model the response probability as
\begin{equation}
\Pr(R=1\mid A=1,Y=y)
=
r(y;\theta,\eta),
\qquad
0<r(y;\theta,\eta)<1,
\label{eq:response-model}
\end{equation}
where \(\eta\in\mathcal H\subseteq\mathbb R^q\) contains parameters specific
to the response mechanism. The defining feature of
\eqref{eq:response-model} is that the oracle response may depend on the same
parameter \(\theta\) that determines the classifier. The realized response or
abstention can therefore itself contain information about the classification
model.

Throughout the main development we assume
\begin{equation}
R\perp Z\mid(Y,A=1),
\label{eq:response-mar}
\end{equation}
and that a returned label is correct. Thus the response mechanism is
feature-dependent but does not depend further on the unobserved class once
\(Y\) is given. Noisy returned labels and class-dependent response mechanisms
are outside the scope of the present paper.

For one sampling unit, the observed data can be represented by
\[
O=(Y,A,AR,ARZ).
\]
Up to factors involving the known acquisition design \(a\), its likelihood is
\begin{equation}
L(\theta,\eta;O)
\propto
p_\theta(Y)
\left[
r(Y;\theta,\eta)^R
\{1-r(Y;\theta,\eta)\}^{1-R}
\tau_Z(Y;\theta)^R
\right]^A.
\label{eq:observed-likelihood}
\end{equation}
Hence
\begin{align}
\ell(\theta,\eta;O)
&=
\log p_\theta(Y)
\nonumber\\
&\quad+
A\left[
R\log r(Y;\theta,\eta)
+(1-R)\log\{1-r(Y;\theta,\eta)\}
+R\log\tau_Z(Y;\theta)
\right].
\label{eq:observed-loglik}
\end{align}

Define the feature and conditional-label scores by
\[
S_Y(Y;\theta)
=
\nabla_\theta\log p_\theta(Y),
\qquad
S_{Z\mid Y}(Y,Z;\theta)
=
\nabla_\theta\log\tau_Z(Y;\theta),
\]
and let
\[
q_\theta(y)
=
\nabla_\theta\operatorname{logit}r(y;\theta,\eta),
\qquad
q_\eta(y)
=
\nabla_\eta\operatorname{logit}r(y;\theta,\eta).
\]
Differentiation of \eqref{eq:observed-loglik} gives
\begin{align}
S_\theta
&=
S_Y
+
A\left[
\{R-r(Y)\}q_\theta(Y)
+
R S_{Z\mid Y}
\right],
\label{eq:score-theta}\\
S_\eta
&=
A\{R-r(Y)\}q_\eta(Y).
\label{eq:score-eta}
\end{align}
Thus a query can generate information through two distinct channels: the
class label when it is returned and the response indicator itself.

Let
\[
J(y)
=
E_\theta
\left[
S_{Z\mid Y}S_{Z\mid Y}^{\top}
\mid Y=y
\right]
\]
denote the conditional Fisher information in a returned class label, and set
\[
w(y)=r(y)\{1-r(y)\},
\qquad
q(y)=
\begin{pmatrix}
q_\theta(y)\\
q_\eta(y)
\end{pmatrix},
\qquad
\widetilde J(y)=
\begin{pmatrix}
J(y)&0\\
0&0
\end{pmatrix}.
\]
The information generated by a query at feature value \(y\) is
\begin{equation}
K(y)
=
r(y)\widetilde J(y)
+
w(y)q(y)q(y)^\top.
\label{eq:query-information}
\end{equation}
The first term is the expected information from the returned label, whereas
the second is the information carried by the response or abstention outcome.

Let
\[
\phi=(\theta^\top,\eta^\top)^\top,
\qquad
\mathcal I_Y=
\begin{pmatrix}
I_Y(\theta)&0\\
0&0
\end{pmatrix},
\]
where
\[
I_Y(\theta)
=
E_\theta\{S_Y(Y;\theta)S_Y(Y;\theta)^\top\}.
\]

\begin{proposition}
\label{prop:information-decomposition}
Suppose that \eqref{eq:acquisition-design}--\eqref{eq:response-mar} hold and
that the usual regularity conditions for Fisher-information calculations are
satisfied. For a fixed acquisition design \(a\), the per-observation Fisher
information for \(\phi\) is
\begin{equation}
\mathcal I(a)
=
\mathcal I_Y
+
E_\theta\!\left[
a(Y)K(Y)
\right],
\label{eq:information-design}
\end{equation}
where \(K(y)\) is defined in \eqref{eq:query-information}. Equivalently,
\begin{equation}
\mathcal I(a)
=
\mathcal I_Y+
E_\theta\!\left[
a(Y)
\left\{
r(Y)\widetilde J(Y)
+
r(Y)\{1-r(Y)\}q(Y)q(Y)^\top
\right\}
\right].
\label{eq:information-design-expanded}
\end{equation}
\end{proposition}

The proof is given in Supplementary
Section~\ref{supp:proof-information-decomposition}.

Proposition~\ref{prop:information-decomposition} separates the two sources of
information created by a label request. If a requested label is returned with
certainty, \(r(y)\equiv1\), then the response term vanishes and
\eqref{eq:information-design-expanded} reduces to the usual information for
selective label acquisition. More generally, if the response mechanism is
parameter-dependent, the statistical value of a query is not determined only
by the probability of obtaining its label: the observed response or
abstention is itself part of the annotation experiment.

Because \(\eta\) is unknown, however, the second term in
\eqref{eq:query-information} cannot be interpreted directly as additional
information about the classifier. The component that remains informative for
\(\theta\) after adjustment for the response-model nuisance parameters is
derived next.

\section{Efficient classifier information and the value of oracle response}
\label{sec:efficient-information}

The information decomposition in
Proposition~\ref{prop:information-decomposition} treats
\(\phi=(\theta^\top,\eta^\top)^\top\) as the full parameter vector. For
classification, however, \(\theta\) is the parameter of interest, whereas
\(\eta\) is a nuisance parameter governing the oracle-response mechanism.
Consequently, the response-information term in
\eqref{eq:information-design-expanded} must be adjusted for uncertainty in
\(\eta\).

Partition the Fisher information according to
\(\phi=(\theta^\top,\eta^\top)^\top\):
\[
\mathcal I(a)
=
\begin{pmatrix}
I_{\theta\theta}(a) & I_{\theta\eta}(a)\\
I_{\eta\theta}(a) & I_{\eta\eta}(a)
\end{pmatrix}.
\]
From \eqref{eq:information-design-expanded},
\begin{align}
I_{\theta\theta}(a)
&=
I_Y(\theta)
+
E_\theta\!\left[
a(Y)r(Y)J(Y)
\right]
+
C_{\theta\theta}(a),
\label{eq:I-theta-theta}\\
I_{\theta\eta}(a)
&=
C_{\theta\eta}(a),
\label{eq:I-theta-eta}\\
I_{\eta\eta}(a)
&=
C_{\eta\eta}(a),
\label{eq:I-eta-eta}
\end{align}
where
\begin{align}
C_{\theta\theta}(a)
&=
E_\theta\!\left[
a(Y)w(Y)q_\theta(Y)q_\theta(Y)^\top
\right],
\label{eq:C-theta-theta}\\
C_{\theta\eta}(a)
&=
E_\theta\!\left[
a(Y)w(Y)q_\theta(Y)q_\eta(Y)^\top
\right],
\label{eq:C-theta-eta}\\
C_{\eta\eta}(a)
&=
E_\theta\!\left[
a(Y)w(Y)q_\eta(Y)q_\eta(Y)^\top
\right].
\label{eq:C-eta-eta}
\end{align}

Assume throughout this section that \(C_{\eta\eta}(a)\) is positive definite.
The efficient Fisher information for \(\theta\), after adjustment for
\(\eta\), is the Schur complement
\begin{equation}
I_{\mathrm{eff}}(a)
=
I_{\theta\theta}(a)
-
I_{\theta\eta}(a)
I_{\eta\eta}(a)^{-1}
I_{\eta\theta}(a).
\label{eq:efficient-information}
\end{equation}
Substitution of \eqref{eq:I-theta-theta}--\eqref{eq:I-eta-eta} yields
\begin{equation}
I_{\mathrm{eff}}(a)
=
I_Y(\theta)
+
E_\theta\!\left[
a(Y)r(Y)J(Y)
\right]
+
I_{R,\mathrm{eff}}(a),
\label{eq:efficient-information-decomposition}
\end{equation}
where
\begin{equation}
I_{R,\mathrm{eff}}(a)
=
C_{\theta\theta}(a)
-
C_{\theta\eta}(a)
C_{\eta\eta}(a)^{-1}
C_{\eta\theta}(a).
\label{eq:efficient-response-information}
\end{equation}
The first two terms in
\eqref{eq:efficient-information-decomposition} are respectively the
information available from the feature distribution and the expected
information from labels that are actually returned. The final term measures
the additional classifier information supplied by the response process after
removing the component attributable to the nuisance parameter \(\eta\).

The following result gives a direct characterization of this contribution.

\begin{proposition}
\label{prop:efficient-response-information}
Suppose that \(C_{\eta\eta}(a)\) is positive definite and define
\begin{equation}
B_a
=
C_{\theta\eta}(a)C_{\eta\eta}(a)^{-1}.
\label{eq:B-a}
\end{equation}
Then
\begin{equation}
I_{R,\mathrm{eff}}(a)
=
E_\theta\!\left[
a(Y)w(Y)
\{q_\theta(Y)-B_aq_\eta(Y)\}
\{q_\theta(Y)-B_aq_\eta(Y)\}^{\top}
\right].
\label{eq:response-projection}
\end{equation}
Consequently,
\[
I_{R,\mathrm{eff}}(a)\succeq0.
\]
Moreover,
\begin{equation}
I_{R,\mathrm{eff}}(a)=0
\quad\Longleftrightarrow\quad
q_\theta(Y)=B_aq_\eta(Y)
\quad
\mu_a\text{-almost surely},
\label{eq:zero-response-information}
\end{equation}
where \(\mu_a\) is the measure with density proportional to
\(a(y)w(y)p_\theta(y)\).
\end{proposition}

The proof is given in Supplementary
Section~\ref{supp:proof-efficient-response-information}.

Proposition~\ref{prop:efficient-response-information} identifies precisely
when observing whether the oracle responds provides information about the
classifier beyond that contained in returned labels. The vector
\(q_\theta(y)\) describes the local sensitivity of the response probability to
the classifier parameter, while \(q_\eta(y)\) describes sensitivity to the
response-specific nuisance parameter. Equation~\eqref{eq:response-projection}
shows that only the component of \(q_\theta(Y)\) that cannot be represented
through the nuisance score contributes efficient information for
classification.

This distinction is important. Parameter dependence of the response
probability alone does not guarantee additional information about
\(\theta\) after nuisance adjustment. If
\eqref{eq:zero-response-information} holds, all classifier-related variation
in the response mechanism is statistically confounded with variation in
\(\eta\), and the response indicator contributes no additional efficient
information. Otherwise,
\[
I_{R,\mathrm{eff}}(a)\neq0,
\]
and observing response or abstention supplies classifier information that
would be discarded by a design criterion based only on the probability of
obtaining a label.

The decomposition also identifies two limiting cases. If the response
mechanism is independent of the classifier parameter, then
\(q_\theta(y)\equiv0\) and hence
\[
I_{R,\mathrm{eff}}(a)=0.
\]
The Bernoulli weight \(w(y)=r(y)\{1-r(y)\}\) approaches zero
as the response probability approaches either endpoint.
If the corresponding logit-score vectors remain bounded,
the pointwise response-information contribution also vanishes.

The positive-definiteness requirement on \(C_{\eta\eta}(a)\) is itself
design-dependent. An acquisition rule may therefore be highly informative
for classification while failing to identify the response mechanism. We
return to this issue when defining the admissible class of acquisition
designs.

\section{Classification risk under informative oracle response}
\label{sec:risk}

The information criterion in Section~\ref{sec:efficient-information} measures
precision in estimating the model parameters, but the objective of label
acquisition is classification. We therefore evaluate an acquisition design
through its leading contribution to the zero-one classification risk.

Let
\[
d_\theta(y)
=
\arg\max_{1\leq k\leq g}\tau_k(y;\theta)
\]
denote the plug-in Bayes rule and define its risk, under the true parameter
\(\theta_0\), by
\[
R(\theta)
=
\Pr_{\theta_0}\{d_\theta(Y)\neq Z\}.
\]
Let
\[
R^*=R(\theta_0)
\]
be the Bayes risk. Under the regular boundary conditions for multiclass
classification, the local excess risk admits the quadratic expansion
\begin{equation}
R(\theta_0+h)-R^*
=
\frac{1}{2}h^\top H_R h
+
o(\|h\|^2),
\qquad h\rightarrow0,
\label{eq:risk-expansion}
\end{equation}
where \(H_R\succeq0\) is the classification-risk curvature matrix at
\(\theta_0\). It depends only on parameter directions that perturb active
Bayes decision boundaries.

The response parameter \(\eta\) does not enter the classifier. Accordingly,
for the full parameter
\[
\phi=(\theta^\top,\eta^\top)^\top,
\]
define the augmented risk-curvature matrix
\begin{equation}
\widetilde H_R
=
\begin{pmatrix}
H_R&0\\
0&0
\end{pmatrix}.
\label{eq:augmented-risk-curvature}
\end{equation}

Suppose that, under a fixed admissible acquisition design \(a\), an efficient
estimator \(\widehat\phi_a\) satisfies
\begin{equation}
\sqrt n(\widehat\phi_a-\phi_0)
\overset{d}{\longrightarrow}
N\{0,\mathcal I(a)^{-1}\},
\label{eq:asymptotic-normality}
\end{equation}
with \(\mathcal I(a)\) nonsingular. Assume additionally that, writing
\(e_{n,a}=\widehat\theta_a-\theta_0\),
\[
nE(e_{n,a}e_{n,a}^{\top})\longrightarrow I_{\mathrm{eff}}(a)^{-1},
\qquad
nE\left|R(\widehat\theta_a)-R^*
-\tfrac12 e_{n,a}^{\top}H_Re_{n,a}\right|\longrightarrow0.
\]
These moment and remainder conditions, together with
\eqref{eq:risk-expansion}, give the leading expected excess risk
\begin{equation}
E\{R(\widehat\theta_a)\}-R^*
=
\frac{1}{2n}
\operatorname{tr}
\left\{
\widetilde H_R\mathcal I(a)^{-1}
\right\}
+
o(n^{-1}).
\label{eq:expected-excess-risk}
\end{equation}
Equivalently, using the block-inverse identity,
\begin{equation}
\operatorname{tr}
\left\{
\widetilde H_R\mathcal I(a)^{-1}
\right\}
=
\operatorname{tr}
\left\{
H_R I_{\mathrm{eff}}(a)^{-1}
\right\},
\label{eq:risk-efficient-equivalence}
\end{equation}
where \(I_{\mathrm{eff}}(a)\) is the nuisance-adjusted information in
\eqref{eq:efficient-information}.

This motivates the classification-risk criterion
\begin{equation}
\Phi(a)
=
\operatorname{tr}
\left\{
\widetilde H_R\mathcal I(a)^{-1}
\right\}.
\label{eq:risk-criterion}
\end{equation}
For a labeling budget \(\rho\in(0,1)\), the oracle acquisition problem is
therefore
\begin{equation}
\min_{a\in\mathcal A_\rho}\Phi(a),
\qquad
\mathcal A_\rho
=
\left\{
a:\ 0\leq a(y)\leq1,\ 
E_{\theta_0}\{a(Y)\}=\rho,\ 
\mathcal I(a)\succ0
\right\}.
\label{eq:oracle-design-problem}
\end{equation}
The nonsingularity condition is substantive in the present setting because
the acquisition design determines not only which labels may be observed but
also which response outcomes are available for identifying the response
mechanism.

Since
\[
\mathcal I(a)
=
\mathcal I_Y+E\{a(Y)K(Y)\}
\]
is affine in \(a\), the criterion in \eqref{eq:risk-criterion} retains the
trace-inverse structure of an optimal-design problem despite the nuisance
adjustment in \(I_{\mathrm{eff}}(a)\).

The local value of querying a feature vector \(y\) is obtained from the
directional derivative of \(\Phi\). Define
\begin{equation}
\psi_a(y)
=
\operatorname{tr}
\left\{
\widetilde H_R
\mathcal I(a)^{-1}
K(y)
\mathcal I(a)^{-1}
\right\}.
\label{eq:query-value}
\end{equation}
Because
\[
K(y)
=
r(y)\widetilde J(y)
+
w(y)q(y)q(y)^\top,
\]
the query value decomposes as
\begin{equation}
\psi_a(y)
=
\psi_a^{L}(y)
+
\psi_a^{R}(y),
\label{eq:query-value-decomposition}
\end{equation}
where
\begin{align}
\psi_a^{L}(y)
&=
r(y)
\operatorname{tr}
\left\{
\widetilde H_R
\mathcal I(a)^{-1}
\widetilde J(y)
\mathcal I(a)^{-1}
\right\},
\label{eq:label-value}\\
\psi_a^{R}(y)
&=
w(y)
q(y)^\top
\mathcal I(a)^{-1}
\widetilde H_R
\mathcal I(a)^{-1}
q(y).
\label{eq:response-value}
\end{align}

The term \(\psi_a^{L}(y)\) is the classification value of the label,
discounted by the probability that the oracle returns it. In contrast,
\(\psi_a^{R}(y)\) is the classification value of observing the response
outcome itself. In particular,
\[
\psi_a^{R}(y)\geq0.
\]
Thus the value of a query is not, in general, obtained by simply multiplying
a label-acquisition score by the probability of response. Such a calculation
retains \(\psi_a^{L}(y)\) but discards \(\psi_a^{R}(y)\).

\begin{proposition}
\label{prop:risk-query-value}
Suppose that \(\mathcal I(a)\succ0\), and let \(h\) be a bounded measurable
perturbation such that \(a_t=a+th\) remains admissible for sufficiently small
\(|t|\). Then
\begin{equation}
\left.
\frac{d}{dt}\Phi(a_t)
\right|_{t=0}
=
-
E_{\theta_0}\!\left\{
h(Y)\psi_a(Y)
\right\},
\label{eq:gateaux-derivative}
\end{equation}
where \(\psi_a\) is given by \eqref{eq:query-value}. Moreover,
\(\psi_a\) admits the decomposition
\eqref{eq:query-value-decomposition}--\eqref{eq:response-value}.
\end{proposition}

The proof is given in Supplementary
Section~\ref{supp:proof-risk-query-value}.

Proposition~\ref{prop:risk-query-value} provides the local principle underlying
response-aware acquisition. A candidate can be valuable because its label is
likely to be returned and informative for the decision boundary, because its
response outcome is itself informative for classifier-relevant parameter
directions, or through both channels. Consequently, two candidates can be
ordered differently by a response-discounted label criterion and by the full
classification-risk criterion. The optimal allocation implied by this
marginal value is characterized in the next section.

\section{Risk-optimal acquisition and response identifiability}
\label{sec:optimal-design}

We now characterize acquisition designs that minimize the leading
classification-risk criterion
\[
\Phi(a)
=
\operatorname{tr}
\left\{
\widetilde H_R\mathcal I(a)^{-1}
\right\}
\]
subject to a fixed query budget. Because the information matrix is affine in
the acquisition design, the resulting problem retains a convex
optimal-design structure even though the response mechanism contains unknown
nuisance parameters.

For a fixed budget \(\rho\in(0,1)\), consider
\begin{equation}
\mathcal A_\rho^\circ
=
\left\{
a:\mathcal Y\rightarrow[0,1]:
E_{\theta_0}\{a(Y)\}=\rho,\ 
\mathcal I(a)\succ0
\right\}.
\label{eq:admissible-design-class}
\end{equation}
We assume that this class is nonempty and that an optimal design exists.

\begin{proposition}
\label{prop:convexity}
On any convex subset of \(\mathcal A_\rho^\circ\), the criterion
\[
a\longmapsto\Phi(a)
\]
is convex.
\end{proposition}

The proof is given in Supplementary
Section~\ref{supp:proof-convexity}.

Convexity, together with the derivative in
Proposition~\ref{prop:risk-query-value}, yields a direct characterization of
the oracle acquisition rule.

\begin{theorem}
\label{thm:optimal-acquisition}
Let \(a^*\in\mathcal A_\rho^\circ\) minimize \(\Phi(a)\), and suppose that
\(a^*\) is an interior point of the positive-definite information domain
relative to the box and budget constraints. Then there exists
\(\lambda\in\mathbb R\) such that, for
\(P_{\theta_0}\)-almost every \(y\),
\begin{equation}
\begin{cases}
\psi_{a^*}(y)>\lambda
&\Longrightarrow\quad a^*(y)=1,\\[3pt]
\psi_{a^*}(y)<\lambda
&\Longrightarrow\quad a^*(y)=0,\\[3pt]
0<a^*(y)<1
&\Longrightarrow\quad \psi_{a^*}(y)=\lambda,
\end{cases}
\label{eq:optimal-threshold-rule}
\end{equation}
where
\[
\psi_{a^*}(y)
=
\psi_{a^*}^{L}(y)+\psi_{a^*}^{R}(y)
\]
is the response-aware classification value defined in
\eqref{eq:query-value-decomposition}. Conversely, any feasible design
satisfying these conditions is globally optimal.
\end{theorem}

The proof is given in Supplementary
Section~\ref{supp:proof-optimal-acquisition}.

Theorem~\ref{thm:optimal-acquisition} shows that the optimal design allocates
the labeling budget according to the total classification value of a query,
rather than according to the value of a potentially returned label alone.
Except on level sets where randomization is required to satisfy the budget,
the oracle rule therefore has a threshold form.

This distinction can alter the ordering of candidate observations. A
response-discounted rule based only on
\(\psi_a^{L}(y)\) favors observations whose returned labels are valuable after
accounting for their probability of response. The full rule instead ranks
observations according to
\[
\psi_a^{L}(y)+\psi_a^{R}(y).
\]
Consequently, an observation with a smaller probability of returning a label
may still have greater total acquisition value if its response outcome is
sufficiently informative about classifier-relevant parameter directions.
The phenomenon does not require response probabilities close to zero:
indeed, the Bernoulli information factor
\(w(y)=r(y)\{1-r(y)\}\) vanishes as \(r(y)\) approaches either zero or one.

\subsection{Response identifiability under selective acquisition}
\label{subsec:response-identifiability}

The preceding optimality result presumes that the information matrix is
nonsingular. In the present problem this requirement depends on the
acquisition design itself. A design that concentrates queries on a restricted
region of the feature space can provide substantial classification
information while failing to identify the response mechanism.

The nuisance-information block is
\begin{equation}
C_{\eta\eta}(a)
=
E_{\theta_0}\!\left[
a(Y)w(Y)q_\eta(Y)q_\eta(Y)^\top
\right].
\label{eq:response-nuisance-information}
\end{equation}

\begin{proposition}
\label{prop:response-identifiability}
For any \(v\in\mathbb R^q\),
\begin{equation}
v^\top C_{\eta\eta}(a)v
=
E_{\theta_0}\!\left[
a(Y)w(Y)
\{v^\top q_\eta(Y)\}^2
\right].
\label{eq:response-identifiability-quadratic}
\end{equation}
Consequently,
\[
C_{\eta\eta}(a)\succ0
\]
if and only if there is no nonzero \(v\in\mathbb R^q\) such that
\begin{equation}
v^\top q_\eta(Y)=0
\qquad
\mu_a\text{-almost surely},
\label{eq:response-rank-condition}
\end{equation}
where \(\mu_a\) is the weighted measure introduced in
Proposition~\ref{prop:efficient-response-information}.
\end{proposition}

The proof is given in Supplementary
Section~\ref{supp:proof-response-identifiability}.

Proposition~\ref{prop:response-identifiability} gives a design-specific rank
condition for estimation of the response nuisance parameter. In particular,
positive query probability alone is insufficient: the queried observations
must exhibit enough variation in the response-model regressors to distinguish
the components of \(\eta\).

For example, consider the response model
\begin{equation}
\operatorname{logit}r(y;\theta,\eta)
=
\eta_0+\eta_1 U(y;\theta),
\label{eq:uncertainty-response-model}
\end{equation}
where \(U(y;\theta)\) is a scalar measure of classification uncertainty. Then
\[
q_\eta(y)
=
\begin{pmatrix}
1\\
U(y;\theta)
\end{pmatrix}.
\]
Provided \(a(Y)w(Y)\) has positive expectation,
\(C_{\eta\eta}(a)\succ0\) if and only if
\begin{equation}
\operatorname{Var}_{\mu_a}\{U(Y;\theta_0)\}>0.
\label{eq:uncertainty-identifiability}
\end{equation}
Thus a design supported only on observations having the same uncertainty
level cannot separately identify the intercept and uncertainty effect in the
response model. In particular, concentrating all queries on a single
decision-boundary uncertainty level can destroy response-mechanism
identifiability.

For theoretical and numerical implementation it is therefore useful to work
with the uniformly regular design class
\begin{equation}
\mathcal A_{\rho,\kappa}
=
\left\{
a:\ 0\leq a(y)\leq1,\ 
E_{\theta_0}\{a(Y)\}=\rho,\ 
\lambda_{\min}\{\mathcal I(a)\}\geq\kappa
\right\},
\qquad \kappa>0.
\label{eq:regular-design-class}
\end{equation}
The lower information bound prevents the acquisition procedure from obtaining
an apparent classification advantage by approaching a design under which
some components of the annotation experiment cease to be identifiable. This
constraint is one of the regularity requirements used in the two-stage
analysis below; it does not by itself establish the estimator conditions
needed for that analysis.

\section{Adaptive response-aware acquisition}
\label{sec:adaptive}

The oracle criterion depends on the unknown classification and response
parameters and on the risk curvature. We first analyze a two-stage
procedure with a pilot-selected design that is held fixed during an
independent second stage. We then distinguish this experiment from the
sequential finite-pool implementation used in the numerical studies.

\subsection{Pilot selection and the second-stage experiment}

Let \(m_n\) be the number of pilot sampling units and let \(n\) be the
number of new second-stage sampling units. Assume
\begin{equation}
m_n\longrightarrow\infty,
\qquad m_n/n\longrightarrow0.
\label{eq:pilot-rate}
\end{equation}
Write \(\mathcal F_{0,n}\) for the sigma-field generated by the pilot
features, query decisions, responses, and returned labels, together with
any auxiliary randomness used to choose the second-stage design. The pilot
uses a design that provides nonsingular information for the parameters to
be estimated. Let \(\widehat\phi_0\) and \(\widehat H_R\) denote the pilot
estimates. The selected acquisition function \(\widehat a_n\) is
\(\mathcal F_{0,n}\)-measurable.

Generate new pairs \((Y_i,Z_i)\), \(i=1,\ldots,n\), independently from
\(p_{\theta_0}\) and independently of \(\mathcal F_{0,n}\). Let
\(U_i,V_i\) be independent uniform random variables on \((0,1)\), independent
across units and of the pilot and the new pairs. Set
\begin{equation}
A_i=\mathbf 1\{U_i\leq\widehat a_n(Y_i)\},
\qquad
A_iR_i=A_i\mathbf 1\{V_i\leq r(Y_i;\phi_0)\}.
\label{eq:second-stage-construction}
\end{equation}
This construction implements the conditional independence of response and
class label assumed in Section~\ref{sec:model}. Define
\(O_i=(Y_i,A_i,A_iR_i,A_iR_iZ_i)\), and let \(P_{\phi,a}\) denote the
one-unit observed-data law under parameter \(\phi\) and a fixed design \(a\).
Then
\begin{equation}
\mathcal L(O_1,\ldots,O_n\mid\mathcal F_{0,n})
=P_{\phi_0,\widehat a_n}^{\otimes n}.
\label{eq:conditional-iid-second-stage}
\end{equation}
Thus the second-stage observations are independent and identically
distributed conditional on the pilot. Unconditionally they generally are
not independent, because they share the random design \(\widehat a_n\).
We do not apply an iid argument to the pooled two-stage sample.

For example, for a fixed design \(a\), the nonzero observed-data densities are
\[
\begin{aligned}
p_{\phi,a}(y,A=0)&=p_\theta(y)\{1-a(y)\},\\
p_{\phi,a}(y,A=1,R=0)&=p_\theta(y)a(y)\{1-r(y;\phi)\},\\
p_{\phi,a}(y,A=1,R=1,Z=z)
&=p_\theta(y)a(y)r(y;\phi)\tau_z(y;\theta).
\end{aligned}
\]
Conditional on the observed pilot, the selected function \(\widehat a_n\)
is held fixed when differentiating the second-stage likelihood. Its
information per sampling unit is therefore \(\mathcal I(\widehat a_n)\)
from Proposition~\ref{prop:information-decomposition}, evaluated at
\(\phi_0\). The returned-label subset need not follow the population
feature law; \eqref{eq:conditional-iid-second-stage} concerns the complete
observed-data units, including unqueried units and queried abstentions.

\subsection{Design selection and oracle value}

Use a fixed, nonempty measurable design class
\(\mathcal D\subseteq\mathcal A_{\rho,\kappa}\), where
\(\mathcal A_{\rho,\kappa}\) is defined in
\eqref{eq:regular-design-class}. One may take the full regular class if
the uniform assumptions below hold; a restricted class may be needed for
empirical estimation. All suprema and selected designs below are assumed
measurable. Define
\begin{equation}
\widehat{\mathcal I}_n(a)
=\widehat{\mathcal I}_{Y,n}
+\widehat E_n\{a(Y)\widehat K_n(Y)\},
\label{eq:plugin-information}
\end{equation}
where
\begin{equation}
\widehat K_n(y)
=\widehat r_n(y)\widetilde J_n(y)
+\widehat r_n(y)\{1-\widehat r_n(y)\}
\widehat q_n(y)\widehat q_n(y)^\top,
\label{eq:plugin-query-information}
\end{equation}
and
\begin{equation}
\widehat\Phi_n(a)
=\operatorname{tr}\{\widehat{\widetilde H}_{R,n}
\widehat{\mathcal I}_n(a)^{-1}\},
\qquad
\widehat{\widetilde H}_{R,n}
=\begin{pmatrix}\widehat H_R&0\\0&0\end{pmatrix}.
\label{eq:plugin-risk-criterion}
\end{equation}
The criterion and every quantity used to choose \(\widehat a_n\) must be
\(\mathcal F_{0,n}\)-measurable. In this experiment, \(\widehat E_n\) may
use pilot features, model-based integration, or an independent reference
feature sample included in \(\mathcal F_{0,n}\). It does not use the features
of the second-stage sample. Positive definiteness of the estimated
information is required on the class over which the criterion is optimized.

Choose a measurable approximate minimizer
\begin{equation}
\widehat a_n\in\mathcal D,
\qquad
\widehat\Phi_n(\widehat a_n)
\leq\inf_{a\in\mathcal D}\widehat\Phi_n(a)+\varepsilon_n,
\qquad 0\leq\varepsilon_n=o_p(1).
\label{eq:adaptive-design}
\end{equation}
Feasibility in the population class \(\mathcal D\) is an assumption.
Using estimated budget and information constraints instead requires a
separate feasibility-approximation argument.

For deterministic \(a\in\mathcal D\), let \(\widehat\phi_{n,a}\) be an
estimator based only on \(n\) iid observations from \(P_{\phi_0,a}\).
It is a measurable function of these observations and \(a\), with no other
pilot dependence. Write \(E_a\) for expectation in this fixed-design
experiment and \(S_a(O;\phi_0)\) for its joint score, given by
\eqref{eq:score-theta}--\eqref{eq:score-eta}. Let
\(\widehat\theta_{n,a}\in\Theta\) be its classification component.

\begin{theorem}
\label{thm:adaptive-oracle}
Suppose that \(a^*\in\arg\min_{a\in\mathcal D}\Phi(a)\) exists,
\eqref{eq:adaptive-design} holds, and
\begin{equation}
\Delta_n:=\sup_{a\in\mathcal D}
|\widehat\Phi_n(a)-\Phi(a)|\overset{p}{\longrightarrow}0.
\label{eq:uniform-criterion-consistency}
\end{equation}
Then
\begin{equation}
0\leq\Phi(\widehat a_n)-\Phi(a^*)
\leq2\Delta_n+\varepsilon_n
\overset{p}{\longrightarrow}0.
\label{eq:oracle-criterion-equivalence}
\end{equation}

For the expected-risk conclusion, additionally assume the independent
second-stage construction
\eqref{eq:second-stage-construction}--\eqref{eq:conditional-iid-second-stage},
the local risk expansion \eqref{eq:risk-expansion}, and the following
uniform fixed-design estimator conditions. First, assume
\begin{equation}
\begin{split}
\sqrt n(\widehat\phi_{n,a}-\phi_0)
&=\mathcal I(a)^{-1}\frac1{\sqrt n}
\sum_{i=1}^n S_a(O_i;\phi_0)+\zeta_{n,a},\\
\sup_{a\in\mathcal D}E_a\|\zeta_{n,a}\|^2&\longrightarrow0.
\end{split}
\label{eq:uniform-l2-linearization}
\end{equation}
Second, for some \(\delta>0\), assume
\begin{equation}
\sup_{n\geq n_0}\sup_{a\in\mathcal D}
E_a\|\sqrt n(\widehat\theta_{n,a}-\theta_0)\|^{2+\delta}<\infty.
\label{eq:uniform-estimator-moment}
\end{equation}
Assume also the usual score identities
\(E_a S_a=0\) and \(E_a(S_aS_a^\top)=\mathcal I(a)\).
Then the second-stage estimator satisfies
\begin{equation}
E\{R(\widehat\theta_{n,\widehat a_n})-R^*
\mid\mathcal F_{0,n}\}
=\frac{\Phi(\widehat a_n)}{2n}+r_n,
\qquad E|r_n|=o(n^{-1}),
\label{eq:conditional-adaptive-risk}
\end{equation}
and consequently
\begin{equation}
E\{R(\widehat\theta_{n,\widehat a_n})\}-R^*
=\frac{\Phi(a^*)}{2n}+o(n^{-1}).
\label{eq:adaptive-risk-equivalence}
\end{equation}
\end{theorem}

The proof is given in Supplementary
Section~\ref{supp:proof-adaptive-oracle}. It first establishes a uniform
expected-risk expansion for deterministic designs and then uses the exact
conditional law \eqref{eq:conditional-iid-second-stage}. In particular,
pointwise asymptotic normality for each fixed design would not suffice:
the selected design varies with the pilot, and convergence in distribution
alone does not justify convergence of expected risk. Conditions
\eqref{eq:uniform-l2-linearization} and
\eqref{eq:uniform-estimator-moment} supply the required uniform second-moment
and remainder control. They are assumptions on the estimator family,
not consequences established here for arbitrary likelihood optimizers.

The information bound gives
\(0\leq\Phi(a)\leq\kappa^{-1}\operatorname{tr}(\widetilde H_R)\)
on \(\mathcal D\). This bound allows the conditional conclusion to be
averaged over the pilot without treating the two stages as iid. The result
does not require a unique oracle design. It is an optimal-value result over
\(\mathcal D\), rather than a proof that the selected design itself
converges to a unique function.

Uniform criterion consistency must also be justified for the chosen
approximation and class. Pilot-parameter consistency alone does not imply
\eqref{eq:uniform-criterion-consistency} for unrestricted measurable designs
when expectations are replaced by empirical averages. A suitable uniform
law of large numbers, an appropriate restricted design class, or direct
uniform control of model-based integration is needed.

The theorem concerns a fixed expected query budget: conditional on the
pilot, the second stage uses \(n\rho\) queries in expectation. Its final
estimator uses only the new observations. If
\(N_n=n+m_n\) counts both stages, \eqref{eq:pilot-rate} gives
\(n/N_n\to1\), so the same risk coefficient can be written with denominator
\(2N_n\). Pilot queries contribute at most \(m_n=o(N_n)\), and the overall
expected query fraction tends to \(\rho\). These accounting statements do
not prove an equivalent result for a pooled estimator. For such an
estimator \(\widetilde\theta_n\), one sufficient additional requirement is
\[
E\|\sqrt n(\widetilde\theta_n-
\widehat\theta_{n,\widehat a_n})\|^2\longrightarrow0,
\qquad
\sup_{n\geq n_0}E\|\sqrt n(\widetilde\theta_n-\theta_0)\|^{2+\delta}<\infty.
\]
The same quadratic-risk argument then transfers the leading coefficient.
The condition \(m_n/n\to0\) alone does not establish these requirements.

\subsection{Sequential finite-pool implementation}

For an available pool \(\{y_1,\ldots,y_N\}\), the plug-in marginal value is
\begin{equation}
\widehat\psi_a(y_i)
=\widehat\psi_a^L(y_i)+\widehat\psi_a^R(y_i).
\label{eq:estimated-query-value}
\end{equation}
A practical implementation updates the fitted classifier, response model,
and information matrices after each batch and selects high-scoring
remaining observations. This uses the query-value decomposition but is a
different sampling experiment from the theorem. Selection without
replacement, exact batch sizes, reuse of pilot observations in estimation,
and repeated model updates generally prevent the conditional product law
\eqref{eq:conditional-iid-second-stage} from holding. Likewise, choosing
the design using the features of the eventual estimation pool does not
satisfy the fresh-sample construction above. Conditioning additionally on
those features leads to a fixed-feature experiment, not the iid
population experiment analyzed here. A separate conditional or adaptive
sampling argument would be needed for that setting.

Accordingly, Theorem~\ref{thm:adaptive-oracle} does not establish oracle
attainment for the sequential finite-pool experiments. Those experiments
assess a practical implementation of the criterion. The pilot provides
initial information about both the classifier and the response mechanism;
a regular pilot design helps avoid beginning acquisition with an
unidentified response model.


\section{Binary logistic illustration}
\label{sec:logistic-illustration}

We now illustrate the preceding theory in a binary classification model for
which the decision boundary, label information, response information, and
classification-risk curvature can all be written explicitly.

Let
\[
Z\in\{0,1\},
\qquad
\Pr_\theta(Z=1\mid Y=y)
=
\tau_\theta(y)
=
\frac{\exp(\alpha+\beta y)}
{1+\exp(\alpha+\beta y)},
\]
with
\[
\theta=(\alpha,\beta)^\top,
\qquad
\beta>0.
\]
For simplicity, suppose that the marginal distribution of \(Y\) does not
depend on \(\theta\), with density \(f\). The Bayes classifier assigns class
1 whenever
\[
\alpha+\beta y>0,
\]
so the decision boundary is
\begin{equation}
b(\theta)=-\frac{\alpha}{\beta}.
\label{eq:logistic-boundary}
\end{equation}

Let
\[
b_0=-\frac{\alpha_0}{\beta_0}
\]
denote the true boundary. Under the usual smoothness condition
\(f(b_0)>0\), the local excess risk satisfies
\begin{equation}
R(\theta_0+h)-R^*
=
\frac{f(b_0)}{4\beta_0}
\left(h_\alpha+b_0h_\beta\right)^2
+
o(\|h\|^2).
\label{eq:logistic-risk-expansion}
\end{equation}
Hence
\begin{equation}
H_R
=
\frac{f(b_0)}{2\beta_0}
v_0v_0^\top,
\qquad
v_0=
\begin{pmatrix}
1\\
b_0
\end{pmatrix}.
\label{eq:logistic-risk-curvature}
\end{equation}
Thus only parameter directions that perturb the decision boundary contribute
to first-order classification risk.

Write
\[
x(y)=
\begin{pmatrix}
1\\
y
\end{pmatrix}.
\]
The conditional class-label information is
\begin{equation}
J(y)
=
\tau(y)\{1-\tau(y)\}x(y)x(y)^\top.
\label{eq:logistic-label-information}
\end{equation}
Because
\[
\tau(y)\{1-\tau(y)\}
\]
is maximized at \(\tau(y)=1/2\), the scalar Bernoulli variance
factor is largest at the Bayes boundary. The full information
matrix also depends on \(x(y)x(y)^\top\), so its magnitude and
classification value need not be largest there.

To model informative oracle response, consider
\begin{equation}
\operatorname{logit}r(y;\theta,\eta)
=
\eta_0+\eta_1U(y;\theta),
\label{eq:entropy-response}
\end{equation}
where
\begin{equation}
U(y;\theta)
=
-\tau(y)\log\tau(y)
-
\{1-\tau(y)\}\log\{1-\tau(y)\}
\label{eq:binary-entropy}
\end{equation}
is the predictive entropy. The case \(\eta_1<0\) corresponds to an oracle
that is less likely to respond at more uncertain observations.

Since
\[
\nabla_\theta\tau(y)
=
\tau(y)\{1-\tau(y)\}x(y)
\]
and
\[
\frac{dU}{d\tau}
=
\log\frac{1-\tau}{\tau},
\]
we obtain
\begin{equation}
\nabla_\theta U(y;\theta)
=
-\{\alpha+\beta y\}
\tau(y)\{1-\tau(y)\}x(y).
\label{eq:entropy-gradient}
\end{equation}
Therefore
\begin{equation}
q_\theta(y)
=
-\eta_1\{\alpha+\beta y\}
\tau(y)\{1-\tau(y)\}x(y),
\label{eq:logistic-q-theta}
\end{equation}
whereas
\begin{equation}
q_\eta(y)
=
\begin{pmatrix}
1\\
U(y;\theta)
\end{pmatrix}.
\label{eq:logistic-q-eta}
\end{equation}

The raw response-information contribution at \(y\) is consequently
\begin{align}
K_R(y)
&=
r(y)\{1-r(y)\}q_\theta(y)q_\theta(y)^\top
\nonumber\\
&=
r(y)\{1-r(y)\}\eta_1^2
\{\alpha+\beta y\}^2
\tau(y)^2\{1-\tau(y)\}^2
x(y)x(y)^\top.
\label{eq:logistic-response-information}
\end{align}
In contrast, the expected returned-label contribution is
\begin{equation}
K_L(y)
=
r(y)
\tau(y)\{1-\tau(y)\}
x(y)x(y)^\top.
\label{eq:logistic-returned-label-information}
\end{equation}

These two sources of information have fundamentally different spatial
geometry. At the Bayes boundary \(y=b(\theta)\),
\[
\alpha+\beta b(\theta)=0,
\]
and hence
\begin{equation}
q_\theta\{b(\theta)\}=0,
\qquad
K_R\{b(\theta)\}=0.
\label{eq:response-zero-boundary}
\end{equation}
Thus, under the entropy-linked response model, the raw classifier information
in the response indicator vanishes exactly at the decision boundary, even
though the scalar Bernoulli variance factor is maximal there.

At the opposite extreme, as \(|\alpha+\beta y|\rightarrow\infty\),
\[
\tau(y)\{1-\tau(y)\}\rightarrow0,
\]
so both the label and response information vanish. Response information is
therefore concentrated away from the exact boundary but not in the extreme
tails. It is naturally associated with intermediate off-boundary regions.

This distinction can change acquisition rankings. A response-discounted
criterion considers only the expected value of the returned label,
represented locally by \(\psi_a^L(y)\). The full response-aware criterion
instead uses
\[
\psi_a(y)
=
\psi_a^L(y)+\psi_a^R(y).
\]
For two candidates \(y_1\) and \(y_2\), it is therefore possible that
\[
\psi_a^L(y_1)>\psi_a^L(y_2),
\]
while
\[
\psi_a(y_1)<\psi_a(y_2).
\]
The reversal occurs when the additional classification value carried by the
response outcome at \(y_2\) more than compensates for its smaller
response-discounted label value.

The example also clarifies why informative response should not be interpreted
as a preference for almost-certain abstention. The Bernoulli factor
\[
r(y)\{1-r(y)\}
\]
vanishes when \(r(y)\) approaches either zero or one. Response information is
largest only where the response probability remains stochastic and its
classifier-related sensitivity is substantial.

The numerical experiments in Section~\ref{sec:simulation} will examine how
this geometry translates into finite-sample acquisition decisions and
classification performance.

\section{Simulation study}
\label{sec:simulation}

We investigated the finite-sample behavior of the response-aware acquisition
criterion in the binary logistic model of Section~\ref{sec:logistic-illustration}.
The simulation was designed to separate two questions. First, we examined
\emph{oracle acquisition}, in which the true model parameters are used to
evaluate candidate query scores. This isolates the statistical effect of the
acquisition criterion itself. Second, we considered an \emph{adaptive}
procedure in which the classifier and response parameters are estimated from
the accumulated query-response data and the acquisition scores are updated
sequentially.

Throughout,
\begin{equation}
Y\sim N(0,1),
\qquad
Z\mid Y=y
\sim
\operatorname{Bernoulli}\{\tau(y)\},
\label{eq:simulation-class-model}
\end{equation}
with
\[
\tau(y)
=
\operatorname{logit}^{-1}(-0.6+2y).
\]
The true Bayes boundary is therefore
\[
b_0=0.3.
\]

The oracle-response mechanism was
\begin{equation}
\operatorname{logit} r(y)
=
\eta_0+\eta_1U(y),
\label{eq:simulation-response-model}
\end{equation}
where \(U(y)\) is the binary predictive entropy defined in
\eqref{eq:binary-entropy}. Three response regimes were considered:
\[
\eta_1\in\{0,-3,-5\},
\]
representing noninformative, moderately informative, and strongly
informative response, respectively. For each value of \(\eta_1\), the
intercept \(\eta_0\) was calibrated so that the marginal response probability
satisfied
\[
E\{r(Y)\}=0.40.
\]
The resulting values were
\[
\eta_0=-0.405,\quad 0.920,\quad 1.786,
\]
with corresponding response probabilities at the Bayes boundary equal to
\(0.400\), \(0.239\), and \(0.157\). Thus the scenarios differ in the
informativeness of the response mechanism rather than simply in the expected
number of returned labels.

For each Monte Carlo replication, an unlabeled pool of size \(3000\) was
generated. Query budgets of
\[
B\in\{300,600,900\}
\]
were considered. These budgets include the common random pilot of
100 queried observations. Subsequent queries were acquired in batches
of 25. The
same potential class labels and oracle-response outcomes were used across
methods within each replication, yielding paired comparisons.

Six acquisition procedures were compared: random sampling, entropy
sampling, ordinary classification-risk-optimal acquisition that treats
labels as always available, response-discounted risk-optimal acquisition,
the proposed response-aware criterion, and a label-only ablation with
full information.

The response-aware criterion uses
\[
\psi_a(y)=\psi_a^L(y)+\psi_a^R(y).
\]
The ablation retains the full query--response information matrix,
including response-model nuisance parameters, but scores candidates
using only \(\psi_a^L(y)\). It therefore differs from
response-discounted acquisition, which uses a label-based information
matrix.

All six procedures used the same full observed-data likelihood after
acquisition. The ablation examines the effect of including the explicit
response-outcome term in sequential query selection while retaining
full-information calculations and joint estimation.
Once the query paths diverge, the procedures naturally have different
accumulated data and fitted parameters.

For adaptive acquisition, the parameters
\[
(\alpha,\beta,\eta_0,\eta_1)
\]
were re-estimated from the accumulated queried data using the likelihood
\[
\ell
=
\sum_{i:A_i=1}
\left[
R_i\log r_i
+
(1-R_i)\log(1-r_i)
+
R_i\left\{
Z_i\log\tau_i
+
(1-Z_i)\log(1-\tau_i)
\right\}
\right].
\]
The principal performance measure was excess classification risk relative to
the true Bayes classifier. We also recorded boundary-estimation error, the
number of returned labels, response rates, and the locations of queried
observations. Results below are based on 500 Monte Carlo replications.
All fits evaluated at the reported budgets returned optimizer
convergence code zero.

Likelihood optimization used box constraints
\[
\alpha\in[-8,8],\qquad
\beta\in[0.05,8],\qquad
\eta_0,\eta_1\in[-12,12].
\]
At each batch, the accumulated information matrices are sums over
all previously queried observations, including the pilot. Ordinary
risk sums \(J(y_i)\), response-discounted risk sums \(r(y_i)J(y_i)\),
and the response-aware and ablation rules both sum \(K(y_i)\).
These quantities are evaluated at the parameters used for scoring
in that batch.
For acquisition calculations, each accumulated information matrix
\(M\) was symmetrized and replaced by
\[
M+\left[
10^{-8}+\max\{0,10^{-8}-\lambda_{\min}(M)\}
\right]I
\]
before inversion. Optimizer convergence codes describe numerical
termination and do not establish global likelihood maximization.

\subsection{Acquisition geometry}
\label{subsec:simulation-geometry}

\begin{figure}[htbp]
    \centering
    \includegraphics[width=\textwidth]{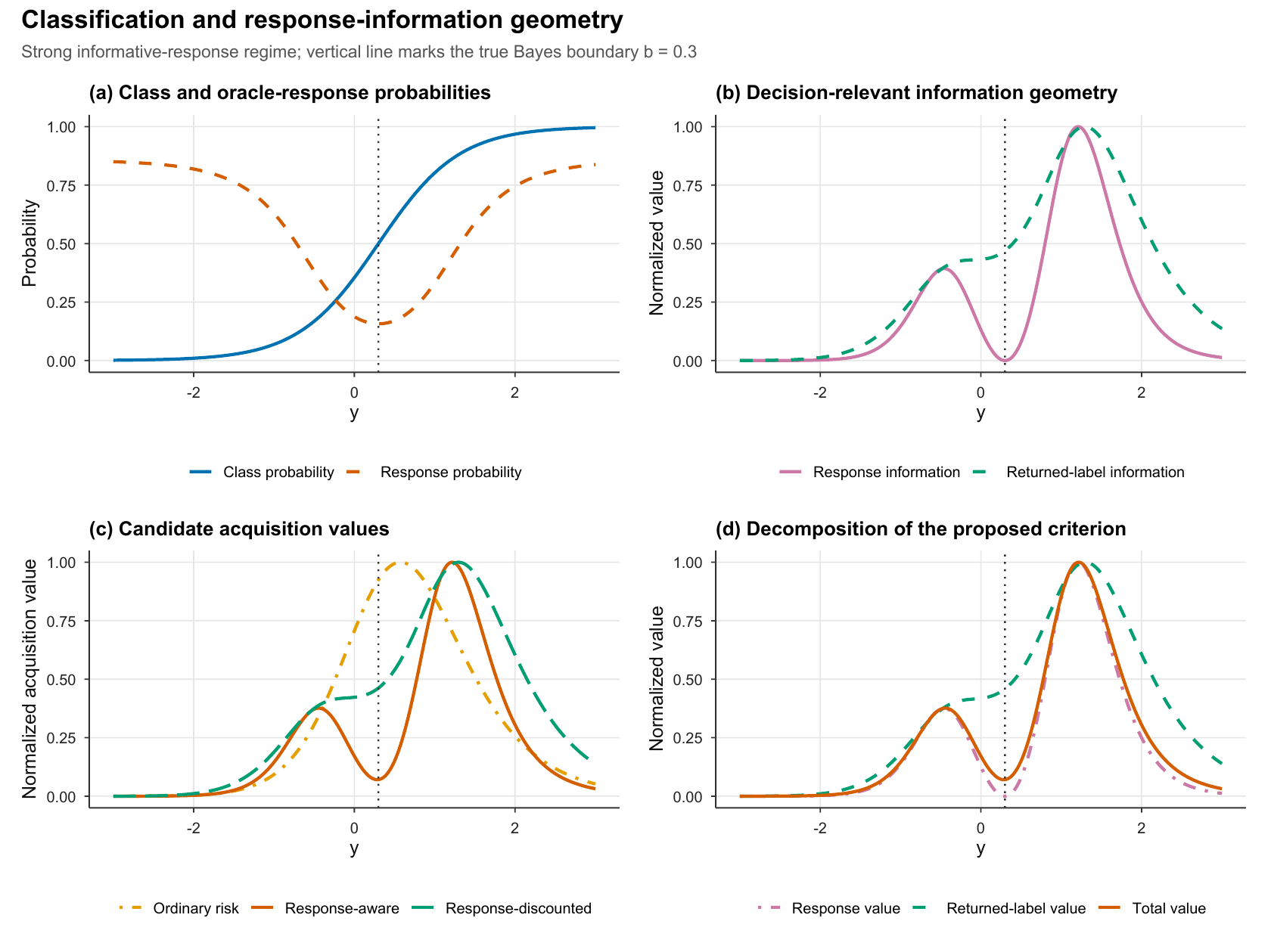}
    \caption{ Information and acquisition geometry in the strongly
    informative-response setting. The vertical dotted line marks the true
    Bayes boundary \(b_0=0.3\). Panel (a) shows the class and response
    probabilities. Panel (b) shows the decision-relevant information from
    returned labels and response outcomes. Panel (c) shows the ordinary,
    response-discounted, and response-aware acquisition scores. Panel (d)
    shows the response-aware score and its two components. The scores in
    panels (c) and (d) are evaluated at the constant reference design
    \(a(y)\equiv 0.5\). Each curve in panels (b)--(d) is divided by its own
    maximum over the plotting grid, so curve heights cannot be used to
    compare magnitudes across curves.}
    \label{fig:simulation-geometry}
\end{figure}

Figure~\ref{fig:simulation-geometry} illustrates the population geometry in
the strongly informative-response regime. The returned-label and response
components have markedly different spatial behavior. The raw classifier-related response-information component vanishes
at the true decision boundary \(b_0=0.3\), as predicted by
\eqref{eq:response-zero-boundary}. The nuisance-adjusted marginal
response value need not vanish there. Numerically, the ordinary risk score is maximized near \(y=0.57\),
whereas the response-discounted and response-aware scores are maximized near
\(y=1.30\) and \(y=1.22\), respectively. The response component itself is
maximized near \(y=1.21\).

The figure therefore demonstrates that accounting for the response indicator
does not merely rescale an uncertainty-based acquisition score. Instead, it
changes the spatial allocation of queries by assigning value to observations
whose response outcome is informative for classifier-relevant parameter
directions.

\subsection{Classification performance}
\label{subsec:simulation-performance}

Table~\ref{tab:adaptive-excess-risk} reports the mean excess classification
risk for the adaptive procedures. Under noninformative response, the
response-aware procedure provides no systematic advantage, as expected. In
the corresponding oracle experiment, the response-aware and
response-discounted procedures coincide exactly for every replication and
budget, providing a direct numerical check of the theoretical reduction
\[
q_\theta(y)=0
\quad\Longrightarrow\quad
\psi_a^R(y)=0.
\]

The picture changes once oracle response is informative. Under moderate
informative response, the response-aware procedure has the smallest mean
excess risk at all three budgets. At \(B=900\), for example, its mean excess
risk is \(0.359\times10^{-3}\), compared with
\(0.463\times10^{-3}\) for the response-discounted rule,
\(0.520\times10^{-3}\) for random acquisition,
\(0.609\times10^{-3}\) for entropy sampling, and
\(0.616\times10^{-3}\) for ordinary risk-optimal acquisition.

The same qualitative pattern appears under strongly informative response.
The proposed procedure attains mean excess risks
\[
0.578,\quad0.249,\quad0.172
\]
in units of \(10^{-3}\) at budgets \(300\), \(600\), and \(900\),
respectively. The corresponding values for the response-discounted method
are
\[
0.656,\quad0.276,\quad0.176.
\]
The gain relative to the response-discounted rule is therefore most visible
at smaller and intermediate budgets and decreases as the available query
budget grows.

The response-aware rule does not dominate in every setting.
In the oracle study, it performs similarly to response-discounted
acquisition in several settings and has slightly higher mean risk in
some. The ablation below examines whether including the explicit
response-outcome term improves acquisition when full information
and joint estimation are retained.

\begin{sidewaystable}[p]
\centering
\caption{Mean excess classification risk multiplied by \(10^3\)
for adaptive acquisition over 500 replications, with Monte Carlo
standard errors in parentheses. The ablation retains full information
but omits the explicit response-outcome acquisition term.
The smallest mean in each row is in bold.}
\label{tab:adaptive-excess-risk}

\small
\setlength{\tabcolsep}{3pt}
\renewcommand{\arraystretch}{1.08}

\begin{tabular}{llcccccc}
\toprule
Response & \(B\)
& Random & Entropy & Ordinary & Discounted
& Ablation & Response-aware \\
\midrule

Noninformative
& 300 & 2.807 (0.169) & 1.890 (0.125)
& \textbf{1.816 (0.106)} & 2.089 (0.125)
& 1.918 (0.111) & 2.138 (0.157) \\
& 600 & 1.495 (0.105) & 0.827 (0.052)
& 0.800 (0.049) & \textbf{0.778 (0.045)}
& 0.825 (0.050) & 0.896 (0.062) \\
& 900 & 0.977 (0.066) & 0.520 (0.034)
& 0.537 (0.035) & \textbf{0.510 (0.032)}
& 0.519 (0.033) & 0.554 (0.035) \\
\addlinespace

Moderate
& 300 & 1.800 (0.114) & 2.218 (0.146)
& 1.927 (0.132) & 1.612 (0.093)
& 1.700 (0.115) & \textbf{1.479 (0.106)} \\
& 600 & 0.763 (0.047) & 1.061 (0.065)
& 0.990 (0.064) & 0.704 (0.043)
& 0.711 (0.053) & \textbf{0.628 (0.045)} \\
& 900 & 0.520 (0.032) & 0.609 (0.037)
& 0.616 (0.038) & 0.463 (0.030)
& 0.467 (0.033) & \textbf{0.359 (0.023)} \\
\addlinespace

Strong
& 300 & 0.926 (0.061) & 1.966 (0.130)
& 1.463 (0.096) & 0.656 (0.040)
& 0.609 (0.039) & \textbf{0.578 (0.039)} \\
& 600 & 0.457 (0.030) & 1.056 (0.068)
& 0.837 (0.056) & 0.276 (0.017)
& 0.275 (0.018) & \textbf{0.249 (0.016)} \\
& 900 & 0.331 (0.024) & 0.634 (0.042)
& 0.604 (0.038) & 0.176 (0.010)
& 0.179 (0.011) & \textbf{0.172 (0.010)} \\
\bottomrule
\end{tabular}
\end{sidewaystable}

\begin{figure}[htbp]
    \centering
    \includegraphics[width=\textwidth]{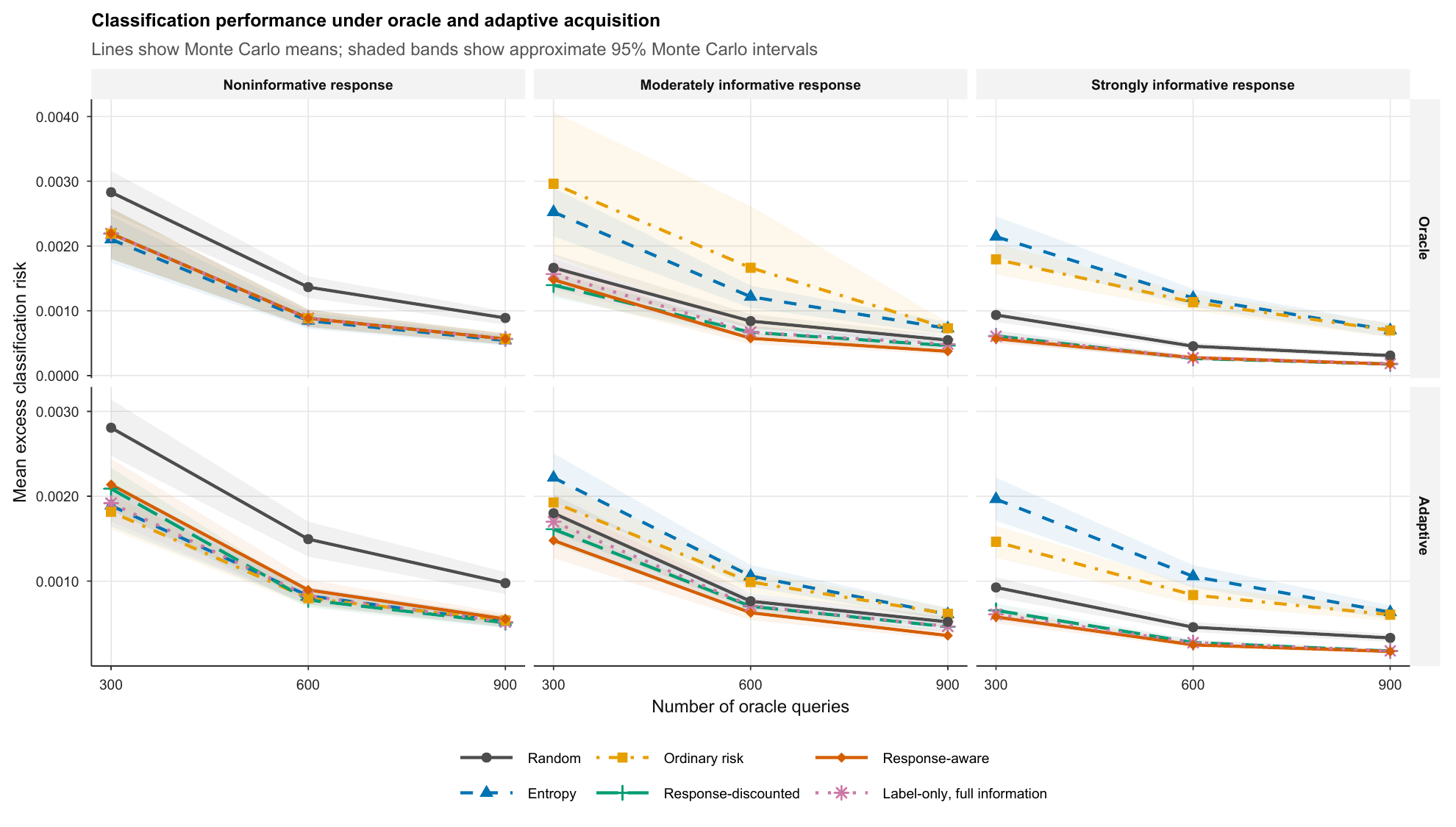}
    \caption{
    Mean excess classification risk as a function of the query budget under
    oracle and adaptive acquisition. Results are based on 500 Monte Carlo
    replications. Shaded regions represent approximate 95\% Monte Carlo
    intervals for the mean.
    }
    \label{fig:simulation-learning-curves}
\end{figure}

Figure~\ref{fig:simulation-learning-curves} gives the corresponding oracle and
adaptive learning curves. The figure includes the label-only, full-information ablation alongside
the five original acquisition procedures. The adaptive response-aware rule also improves
substantially over random acquisition, particularly under strong informative
response. At \(B=900\), its excess risk is approximately \(48\%\) lower than
that of random acquisition in the strong-response setting and approximately
\(31\%\) lower in the moderate-response setting.

\subsection{Contribution of the response-outcome term}
\label{subsec:response-term-ablation}

The ablation retains the full information matrix and joint likelihood,
but omits \(\psi_a^R\) from candidate scores.
Supplementary Table~\ref{tab:supp-response-term-ablation} reports
paired excess-risk differences, defined as the full criterion minus
the ablation.

Under moderately informative response and \(B=900\), the adaptive
procedure reduced mean excess risk from \(0.4666\times10^{-3}\)
to \(0.3589\times10^{-3}\).
The paired difference was \(-0.1077\times10^{-3}\), with a pointwise
95\% Monte Carlo \(t\) interval of
\([-0.1726,-0.0428]\times10^{-3}\) and Holm-adjusted \(p=0.0107\).
The oracle difference was \(-0.1107\times10^{-3}\), with interval
\([-0.1862,-0.0352]\times10^{-3}\) and adjusted \(p=0.0373\).
Holm adjustments cover the nine comparisons within each study.

Relative to the ablation's mean excess risk, both reductions were
approximately \(23\%\). In absolute terms, the adaptive reduction
was approximately \(0.0108\) percentage points of classification error.
The full adaptive criterion had lower mean excess risk in all six
informative-response settings, although the other five intervals
included zero.

Under noninformative response, the oracle procedures had identical
query paths in every saved trace and identical recorded outcomes
across all 500 replications. The adaptive procedures need not coincide:
estimation error can give the fitted response model a nonzero
classifier-dependent component.

The evidence for an additional benefit is therefore confined to the
moderate-response setting at the largest budget. The unresolved
comparisons do not establish equivalence, and the results do not
support a general advantage from increasing response dependence.

\subsection{Where the procedures query}
\label{subsec:simulation-query-location}

\begin{figure}[htbp]
    \centering
    \includegraphics[width=\textwidth]{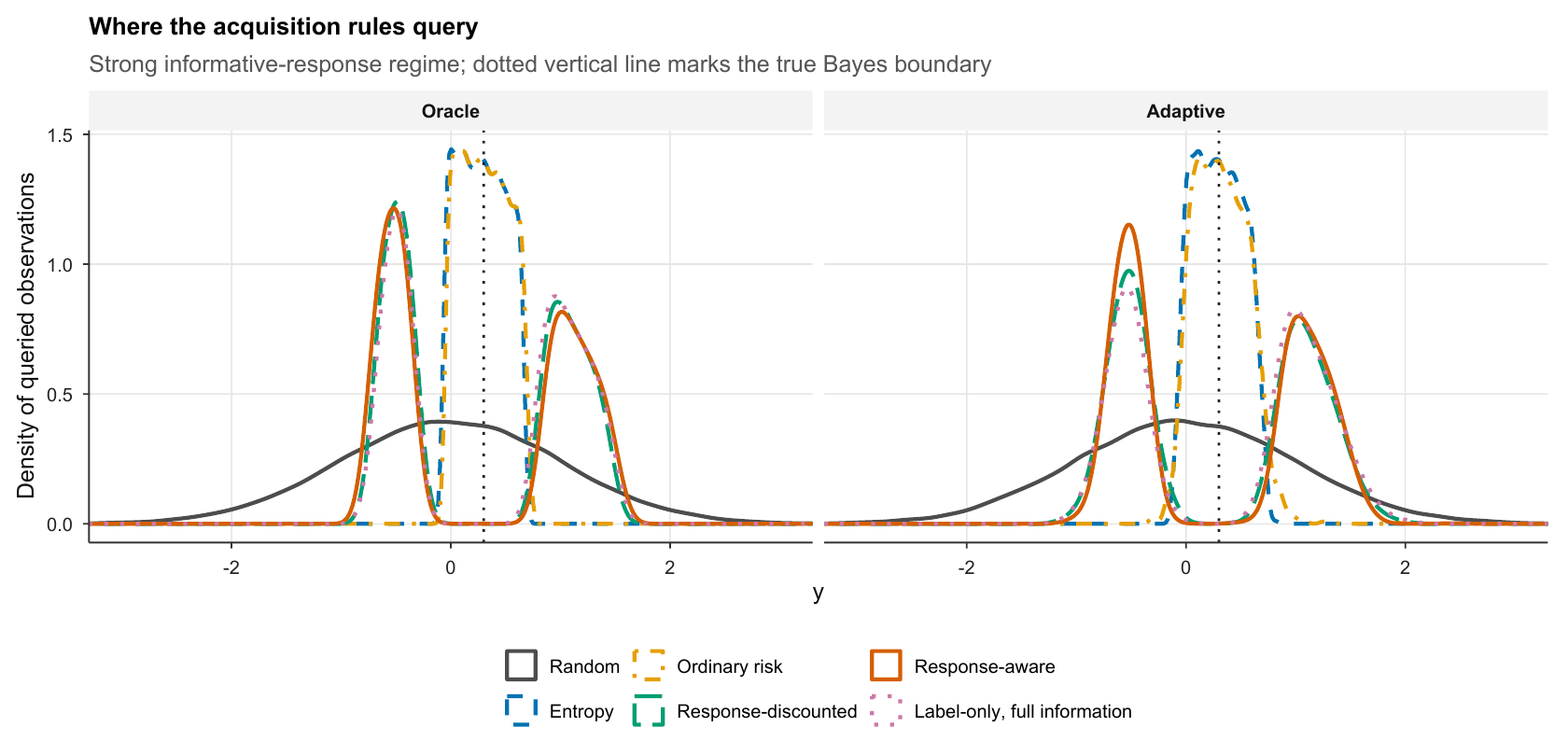}
    \caption{
    Empirical distribution of queried feature values in the strongly
    informative-response setting. The dotted vertical line marks the true
    Bayes boundary. Entropy and ordinary-risk sampling concentrate near the
    boundary, whereas response-discounted and response-aware acquisition
    place substantial query mass in informative off-boundary regions.
    }
    \label{fig:simulation-query-locations}
\end{figure}

The risk differences are accompanied by substantial changes in the locations
of the queried observations. Figure~\ref{fig:simulation-query-locations}
shows the empirical distribution of queried feature values in the strongly
informative-response regime. Entropy and ordinary-risk acquisition place most
of their queries close to the Bayes boundary, where classification
uncertainty is largest. In contrast, both response-discounted and
response-aware acquisition shift substantial mass to off-boundary regions.

For the adaptive procedures at \(B=900\), the mean absolute distance from the
true boundary was \(0.26\), \(0.27\), \(0.83\), and \(0.84\) for entropy,
ordinary-risk, response-discounted, and response-aware acquisition,
respectively. Thus, unlike uncertainty-based rules, the response-aware design
places substantial query mass away from the decision boundary, reflecting the
joint value of returned labels and response information.
The distinction is also visible in the realized response rates. Under strong
informative response and \(B=900\), entropy and ordinary-risk acquisition
obtain responses for only about \(20\%\) of queried observations, whereas
the response-discounted and response-aware procedures obtain response rates
of approximately \(40\%\). The proposed acquisition rule therefore avoids
spending most of its budget in regions where the oracle almost always
abstains, while retaining the additional information carried by the
response outcome itself.

\section{Real-data application}
\label{sec:realdata}

We evaluate the acquisition rules using the Consumer Complaints data from
the AL-RCTA collection of \citet{totakura2026crowd}, which contains
crowd-sourced annotations with explicit annotator abstentions. The analysis uses 2,999 complaints presented
to each of ten human annotators, with the response indicator defined by
\[
R_{ij}
=
\mathbb{I}\{\text{annotator }j\text{ returned a nonzero annotation for
complaint }i\}.
\]
Empirical response rates varied substantially across annotators, from
approximately \(0.75\) to \(0.97\). Response was also related to
classification difficulty: in a preliminary out-of-fold analysis, the
response rate decreased from \(0.972\) in the lowest predictive-entropy
decile to \(0.827\) in the highest.

The theoretical model assumes that a returned label is correct, whereas the
crowd annotations may themselves contain labeling errors. We therefore retain
the observed human response indicator \(R_{ij}\), but, when \(R_{ij}=1\),
reveal the benchmark product category rather than the annotator's reported
class. This preserves the genuine human abstention pattern while isolating
informative response from label noise. Each annotator is treated as a
separate oracle.

Complaint narratives were represented by TF--IDF features followed by a
12-dimensional truncated singular-value decomposition. One of the 3,000 shared complaints was excluded because its narrative
was missing. TF--IDF, singular-value decomposition, feature scaling,
and narrative-length scaling were constructed using all 2,999
narratives before generating the splits. The analysis therefore uses
a transductive feature representation: test narratives contribute to
unsupervised preprocessing, but test labels are withheld from fitting
and acquisition. A multinomial
logistic classifier was fitted jointly with
\[
\operatorname{logit}\{r_j(y;\theta,\eta_j)\}
=
\eta_{0j}+\eta_{1j}U_\theta(y)+\eta_{2j}L(y),
\]
where \(U_\theta(y)\) is normalized predictive entropy and \(L(y)\) is
standardized log narrative length. We compared random, entropy, ordinary
classification-risk, response-discounted risk, and response-aware
acquisition. The classification task comprised six product categories.
Across 30 replications, all methods shared a stratified split with
approximately 70\% of observations allocated to training and 30\% to
testing, a common pilot sample of 300 training observations, and the
same realized oracle responses. Labels were subsequently queried in batches of 50 at
additional budgets \(B\in\{150,300,450,600\}\). All methods used the same penalized joint likelihood, with quadratic
penalty coefficients 1 for classifier slopes and 0.5 for response-model
slopes; intercepts were unpenalized. Risk-based acquisition used an
empirical kernel approximation to the active-boundary curvature,
weighted by the mean fitted probability of the two leading classes.
The Gaussian-kernel bandwidth was the larger of 0.20 and the empirical
25th percentile of their absolute log-probability gaps. The curvature
matrix was trace-normalized. Information matrices used diagonal regularization of \(10^{-3}\).
Before inversion, each eigenvalue of the symmetrized matrix was
replaced by the larger of its original value and
\(\max\{10^{-8}\lambda_{\max},10^{-10}\}\), where
\(\lambda_{\max}\) is the largest eigenvalue. For ordinary-risk and response-discounted acquisition, the accumulated
information matrix uses conditional classifier information from labels
actually returned. Response-aware acquisition instead uses the fitted
full query--response information matrix, including the response-model
nuisance parameters. Thus, the response-discounted baseline is not
obtained simply by removing \(\psi_a^R\) from the response-aware score
while retaining the same information matrix. This experiment compares
the complete acquisition rules rather than isolating the contribution
of that additive term. Each batch selected the 50 highest-scoring available observations
before refitting. This sequential implementation illustrates the
criterion; it is not a numerical verification of the two-stage
oracle-attainment theorem.

\begin{table}[t]
\centering
\caption{Held-out classification error in the Consumer Complaints
application. Entries are means over 30 replications after averaging over
the ten human oracles; parentheses contain Monte Carlo standard errors.
The query budget excludes the common pilot sample of size 300.}
\label{tab:consumer-results}
\small
\setlength{\tabcolsep}{5pt}
\begin{tabular}{lcccc}
\toprule
& \multicolumn{4}{c}{Additional query budget \(B\)}\\
\cmidrule(lr){2-5}
Acquisition rule & 150 & 300 & 450 & 600\\
\midrule
Random
& \(0.260\;(0.0018)\)
& \(0.254\;(0.0017)\)
& \(0.250\;(0.0017)\)
& \(0.247\;(0.0016)\)\\
Entropy
& \(\mathbf{0.246}\;(0.0022)\)
& \(\mathbf{0.241}\;(0.0017)\)
& \(0.238\;(0.0015)\)
& \(0.235\;(0.0015)\)\\
Ordinary risk
& \(0.258\;(0.0019)\)
& \(0.244\;(0.0015)\)
& \(0.237\;(0.0016)\)
& \(0.232\;(0.0016)\)\\
Response-discounted
& \(0.258\;(0.0019)\)
& \(0.244\;(0.0015)\)
& \(0.237\;(0.0016)\)
& \(0.232\;(0.0015)\)\\
Response-aware
& \(0.258\;(0.0019)\)
& \(0.243\;(0.0016)\)
& \(\mathbf{0.235}\;(0.0017)\)
& \(\mathbf{0.231}\;(0.0016)\)\\
\bottomrule
\end{tabular}
\end{table}

Entropy sampling has the lowest mean test error at the two smallest
budgets, whereas response-aware acquisition has the lowest mean error
at \(B=450\) and \(B=600\). The mean paired differences in test error
(response-aware minus response-discounted) are
\[
-0.000052,\quad -0.000892,\quad -0.001152,\quad -0.001152
\]
at \(B=150,300,450,600\), respectively. Pointwise 95\% Monte Carlo
\(t\) intervals exclude zero at the three larger budgets; at
\(B=600\), the interval is \([-0.001679,-0.000624]\).
After Holm correction across the 16 budget--comparator comparisons,
the advantages over response-discounted acquisition remain significant
at \(B=450\) and \(B=600\), with adjusted \(p\)-values \(0.0202\)
and \(0.00111\), respectively. Response-aware acquisition is worse
than entropy sampling at \(B=150\); its advantage over entropy
at \(B=600\) does not remain significant under this correction.
These intervals quantify variability across repeated splits and
acquisition experiments for the fixed dataset and ten annotators,
rather than uncertainty over a population of datasets or annotators.

\begin{figure}[htbp]
\centering
\includegraphics[width=\textwidth]{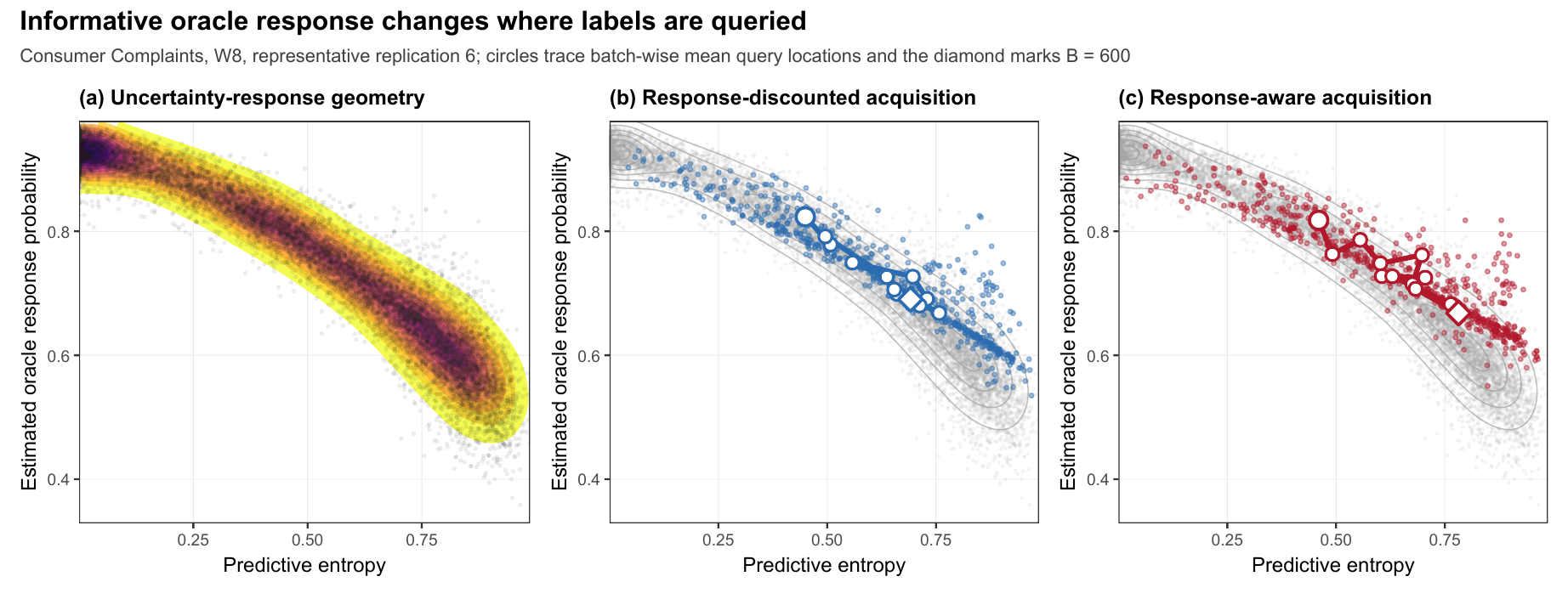}
\caption{ Uncertainty--response acquisition geometry for the Consumer
Complaints application using worker W8, who had the largest observed
abstention fraction. Panel (a) pools random-query locations across
30 replications. Panels (b) and (c) show response-discounted and
response-aware acquisition in replication 6. Coordinates are
predictive entropy and response probability evaluated under each
method's fitted model at the time of querying. Open circles trace
batch-wise mean query locations; the diamond marks the final batch
at additional budget \(B=600\). Replication 6 was selected because
its paired response-aware minus response-discounted test-error
difference was closest to the median across the 30 W8 replications.
The panels describe evolving fitted coordinates, rather than a
common fixed-model representation.}
\label{fig:consumer-geometry}
\end{figure}

Figure~\ref{fig:consumer-geometry} shows a negative association
between fitted predictive entropy and response probability in the
pooled random-query reference cloud. The two risk-based procedures
follow distinct, nonmonotone trajectories in their evolving fitted
coordinates. Their first-batch mean locations are similar.
In the final batch, response-aware acquisition has mean entropy
\(0.781\) and fitted response probability \(0.668\), compared with
\(0.691\) and \(0.690\), respectively, for response-discounted
acquisition. These trajectories illustrate differences in sequential
querying and model updating; they do not by themselves identify
which information component causes the performance difference.

The aggregate improvement is heterogeneous across human oracles.
At \(B=600\), the largest mean improvement over response-discounted
acquisition occurs for W4, with paired difference \(-0.00483\) and
95\% Monte Carlo \(t\) interval \([-0.00768,-0.00198]\).
Only W4 remains significant after Holm correction across the ten
worker-specific comparisons, with adjusted \(p=0.0167\).
The other worker-specific comparisons are statistically unresolved
after this correction.
\section{Discussion and conclusions}
\label{sec:discussion}

This paper extends classification-risk-optimal label acquisition to
an oracle that may abstain and whose response probability shares
parameters with the classifier. The response outcome can then supply
information even when no label is returned. After adjustment for
response-model nuisance parameters, this contribution is
\[
I_{R,\mathrm{eff}}(a)
=
E\!\left[
a(Y)w(Y)
\{q_\theta(Y)-B_aq_\eta(Y)\}
\{q_\theta(Y)-B_aq_\eta(Y)\}^{\top}
\right].
\]
It is positive semidefinite and vanishes when the classifier-related
response score is fully explained by the nuisance response score
under the design-weighted measure. Parameter dependence alone is
therefore insufficient for response outcomes to improve classification.
The remaining information must also act in directions relevant to
the decision boundary.

Combining the information from returned labels and response outcomes
with the local curvature of zero--one risk gives the marginal query
value
\[
\psi_a(y)=\psi_a^L(y)+\psi_a^R(y).
\]
The associated oracle allocation retains the convex trace-inverse
structure of the underlying design problem and admits a threshold
characterization under a fixed query budget. Response identifiability
places an additional restriction on the allocation: queries must
provide enough variation to estimate the response mechanism.

The simulations show how accounting for response changes query
selection. In the strongly informative-response setting,
response-aware and response-discounted acquisition both place
substantial query mass away from the Bayes boundary, where response
is more likely. Entropy and ordinary classification-risk acquisition
remain more concentrated near the boundary. Under noninformative
response, the response-aware and response-discounted oracle procedures
coincide.

The ablation examines the additional effect of including the explicit
response-outcome term while retaining full information and joint
estimation. Under moderately informative response at \(B=900\), the
full criterion reduced mean excess risk in both the oracle and
adaptive studies, with significance after the stated within-study
Holm corrections. The other comparisons were statistically
unresolved. The absolute improvement was small, and the results do
not establish a general advantage across response mechanisms and
budgets. Stronger response dependence alone does not imply a larger
acquisition gain.

The Consumer Complaints replay provides evidence using observed human
abstentions. Response-aware acquisition had the lowest mean test error
at the two largest budgets, with improvements over
response-discounted acquisition remaining significant after Holm
correction at \(B=450\) and \(B=600\). Entropy sampling performed
better at the smallest budget. The improvements were small and varied
across annotators. This comparison evaluates complete acquisition
rules with different information matrices; it does not isolate the
contribution of the explicit response-outcome term.

The theory is local and parametric and relies on correct specification
of the classification and response models. An association between
response and predictive difficulty can motivate a shared-parameter
model, but does not establish that the assumed link is correct.
Performance under misspecification remains unresolved. The two-stage
result applies to a pilot-selected design held fixed on fresh
second-stage observations. It relies on uniform criterion consistency,
mean-square asymptotic linearity, and a higher-moment bound.
It does not establish these estimator conditions, or oracle attainment,
for the sequential finite-pool implementation.

We assume that returned labels are correct and that
\[
R\perp Z\mid(Y,A=1).
\]
The replay therefore preserves human response patterns while revealing
benchmark labels upon response. Using actual worker labels would
require accounting for annotation error and, potentially,
class-dependent response. Joint allocation across annotators with
different costs, response mechanisms, and accuracies would require
further development. 

The main contribution is a classification-risk acquisition criterion
that accounts for information in oracle response outcomes after
nuisance adjustment. The numerical studies support its usefulness
in some settings, while showing that the gains depend on the
response mechanism and query budget.


\section*{Statements and Declarations}

\subsection*{Code and data availability}

Code and reproducibility instructions for the simulation
studies, the Consumer Complaints analysis, and figure
generation are available in the
\href{https://github.com/fariborz-setoudeh/risk-optimal-active-learning-informative-abstention}
{Risk-optimal active learning with informative abstention repository}.
The Consumer Complaints data are obtained from the
original AL-RCTA source cited in the manuscript.

\subsection*{CRediT authorship contribution statement}

\noindent
\textbf{F. Setoudehtazangi:} Conceptualization,
Methodology, Formal analysis, Software, Investigation,
Visualization, Writing -- original draft.
\textbf{G.J. McLachlan:} Methodology, Supervision,
Writing -- review \& editing.

\subsection*{Funding}

This research received no specific grant from funding
agencies in the public, commercial, or not-for-profit
sectors.

\subsection*{Competing interests}

The authors declare no competing interests.

\subsection*{Ethical approval}

This study involved secondary analysis of an existing
dataset and did not involve the recruitment of human
participants or collection of new human-subject data
by the authors.

\setlength{\bibsep}{0.75em}

\clearpage

\begin{center}
{\Large\bfseries Supplementary Material for}\\[0.5em]
{\Large\bfseries Risk-optimal active learning with informative abstention}
\end{center}

\vspace{1.5em}


\setcounter{section}{0}
\setcounter{subsection}{0}
\setcounter{equation}{0}
\setcounter{table}{0}
\setcounter{figure}{0}
\setcounter{theorem}{0}

\renewcommand{\thesection}{S\arabic{section}}
\renewcommand{\thesubsection}{\thesection.\arabic{subsection}}
\renewcommand{\theequation}{\thesection.\arabic{equation}}
\renewcommand{\thetable}{S\arabic{table}}
\renewcommand{\thefigure}{S\arabic{figure}}

\renewcommand{\theHsection}{supp.\arabic{section}}
\renewcommand{\theHsubsection}{supp.\arabic{section}.\arabic{subsection}}
\renewcommand{\theHequation}{supp.\arabic{section}.\arabic{equation}}
\renewcommand{\theHtable}{supp.\arabic{table}}
\renewcommand{\theHfigure}{supp.\arabic{figure}}

\section{Proofs}
\label{supp:proofs}

\subsection{Proof of Proposition~\ref{prop:information-decomposition}}
\label{supp:proof-information-decomposition}

For one observation, write
\[
r(Y)=r(Y;\theta,\eta).
\]
From \eqref{eq:score-theta}--\eqref{eq:score-eta}, the score for
\(\phi=(\theta^\top,\eta^\top)^\top\) can be written as
\begin{equation}
S_\phi
=
\begin{pmatrix}
S_Y\\
0
\end{pmatrix}
+
A
\left[
(R-r)q
+
R\widetilde S_{Z\mid Y}
\right],
\label{eq:supp-full-score}
\end{equation}
where
\[
q=
\begin{pmatrix}
q_\theta\\
q_\eta
\end{pmatrix},
\qquad
\widetilde S_{Z\mid Y}
=
\begin{pmatrix}
S_{Z\mid Y}\\
0
\end{pmatrix}.
\]

Conditional on \(Y=y\), the acquisition indicator satisfies
\[
E(A\mid Y=y)=a(y).
\]
Moreover, by \eqref{eq:response-model},
\[
E(R-r\mid Y=y,A=1)=0
\]
and
\[
E\{(R-r)^2\mid Y=y,A=1\}
=
r(y)\{1-r(y)\}
=
w(y).
\]
Since \(S_{Z\mid Y}\) is a conditional likelihood score,
\[
E(S_{Z\mid Y}\mid Y=y)=0.
\]
Assumption \eqref{eq:response-mar} further implies that, conditional on
\(Y=y\) and \(A=1\), the response indicator \(R\) and the class label \(Z\)
are independent. It follows that
\begin{align}
&E\left[
(R-r)R\widetilde S_{Z\mid Y}
\mid Y=y,A=1
\right]
\nonumber\\
&\qquad=
E\{(R-r)R\mid Y=y,A=1\}
E\left(
\widetilde S_{Z\mid Y}\mid Y=y
\right)
=0.
\label{eq:supp-cross-zero}
\end{align}
Thus the score contribution from the response indicator is orthogonal,
conditional on \(Y\), to the score contribution from the returned class
label.

The conditional second moment of the returned-label term is
\begin{align}
E\left[
R\widetilde S_{Z\mid Y}
\widetilde S_{Z\mid Y}^{\top}
\mid Y=y,A=1
\right]
&=
r(y)
E\left[
\widetilde S_{Z\mid Y}
\widetilde S_{Z\mid Y}^{\top}
\mid Y=y
\right]
\nonumber\\
&=
r(y)\widetilde J(y),
\label{eq:supp-label-info}
\end{align}
where
\[
\widetilde J(y)
=
\begin{pmatrix}
J(y)&0\\
0&0
\end{pmatrix}.
\]
Similarly,
\begin{equation}
E\left[
(R-r)^2qq^\top
\mid Y=y,A=1
\right]
=
w(y)q(y)q(y)^\top.
\label{eq:supp-response-info}
\end{equation}
Combining \eqref{eq:supp-cross-zero}--\eqref{eq:supp-response-info} gives the
conditional Fisher information generated by a query:
\begin{equation}
K(y)
=
r(y)\widetilde J(y)
+
w(y)q(y)q(y)^\top.
\label{eq:supp-query-information}
\end{equation}

It remains to combine the query contribution with the information in the
feature vector. The feature score satisfies
\[
E(S_Y)=0.
\]
Conditional on \(Y\), the mean of the query-score contribution in
\eqref{eq:supp-full-score} is zero:
\begin{align*}
&E\left[
(R-r)q
+
R\widetilde S_{Z\mid Y}
\mid Y,A=1
\right]
\\
&\qquad=
q\,E(R-r\mid Y,A=1)
+
E(R\mid Y,A=1)
E(\widetilde S_{Z\mid Y}\mid Y)
=0.
\end{align*}
Hence the feature score is orthogonal to the query-score contribution. Since
\(A\) is generated according to the known parameter-free design
\eqref{eq:acquisition-design},
\[
E\left[
A K(Y)\mid Y
\right]
=
a(Y)K(Y).
\]
Therefore
\begin{align*}
\mathcal I(a)
&=
E(S_\phi S_\phi^\top)
\\
&=
\begin{pmatrix}
I_Y(\theta)&0\\
0&0
\end{pmatrix}
+
E_\theta\{a(Y)K(Y)\}
\\
&=
\mathcal I_Y
+
E_\theta\{a(Y)K(Y)\},
\end{align*}
which is \eqref{eq:information-design}. Substitution of
\eqref{eq:supp-query-information} yields
\eqref{eq:information-design-expanded}, completing the proof.
\hfill\(\square\)

\subsection{Proof of Proposition~\ref{prop:efficient-response-information}}
\label{supp:proof-efficient-response-information}

For notational simplicity, write
\[
a=a(Y),\qquad
w=w(Y),\qquad
q_\theta=q_\theta(Y),\qquad
q_\eta=q_\eta(Y).
\]
By \eqref{eq:efficient-response-information},
\[
I_{R,\mathrm{eff}}(a)
=
C_{\theta\theta}
-
C_{\theta\eta}C_{\eta\eta}^{-1}C_{\eta\theta}.
\]
Let
\[
B_a=C_{\theta\eta}C_{\eta\eta}^{-1}.
\]
Since \(C_{\eta\eta}\) is symmetric and positive definite,
\[
B_aC_{\eta\eta}
=
C_{\theta\eta},
\qquad
C_{\eta\eta}B_a^\top
=
C_{\eta\theta}.
\]
Consider
\[
E_\theta\!\left[
aw
(q_\theta-B_aq_\eta)
(q_\theta-B_aq_\eta)^\top
\right].
\]
Expanding the outer product and using
\eqref{eq:C-theta-theta}--\eqref{eq:C-eta-eta} gives
\begin{align*}
&E_\theta\!\left[
aw
(q_\theta-B_aq_\eta)
(q_\theta-B_aq_\eta)^\top
\right]
\\
&\quad=
C_{\theta\theta}
-
B_aC_{\eta\theta}
-
C_{\theta\eta}B_a^\top
+
B_aC_{\eta\eta}B_a^\top.
\end{align*}
Because
\[
B_aC_{\eta\theta}
=
C_{\theta\eta}C_{\eta\eta}^{-1}C_{\eta\theta},
\]
and
\[
B_aC_{\eta\eta}B_a^\top
=
C_{\theta\eta}C_{\eta\eta}^{-1}C_{\eta\theta},
\]
the preceding expression reduces to
\[
C_{\theta\theta}
-
C_{\theta\eta}C_{\eta\eta}^{-1}C_{\eta\theta}
=
I_{R,\mathrm{eff}}(a),
\]
which proves \eqref{eq:response-projection}.

For any \(v\in\mathbb R^d\),
\begin{align*}
v^\top I_{R,\mathrm{eff}}(a)v
&=
E_\theta\!\left[
aw
\left\{
v^\top(q_\theta-B_aq_\eta)
\right\}^2
\right]
\geq0.
\end{align*}
Hence
\[
I_{R,\mathrm{eff}}(a)\succeq0.
\]

It remains to characterize when the matrix is zero. If
\[
q_\theta(Y)=B_aq_\eta(Y)
\]
\(\mu_a\)-almost surely, then
\eqref{eq:response-projection} immediately gives
\(I_{R,\mathrm{eff}}(a)=0\).

Conversely, suppose that \(I_{R,\mathrm{eff}}(a)=0\). For every
\(v\in\mathbb R^d\),
\[
E_\theta\!\left[
a(Y)w(Y)
\left\{
v^\top[q_\theta(Y)-B_aq_\eta(Y)]
\right\}^2
\right]
=0.
\]
The integrand is nonnegative, and therefore
\[
v^\top[q_\theta(Y)-B_aq_\eta(Y)]=0
\]
\(\mu_a\)-almost surely for every \(v\in\mathbb R^d\). Taking \(v\) successively
to be the elements of the canonical basis of \(\mathbb R^d\) yields
\[
q_\theta(Y)-B_aq_\eta(Y)=0
\]
\(\mu_a\)-almost surely. This establishes
\eqref{eq:zero-response-information} and completes the proof.
\hfill\(\square\)

\subsection{Proof of Proposition~\ref{prop:risk-query-value}}
\label{supp:proof-risk-query-value}

Let
\[
a_t(y)=a(y)+t h(y)
\]
for \(t\) in a neighborhood of zero such that \(a_t\) remains admissible.
By Proposition~\ref{prop:information-decomposition},
\[
\mathcal I(a_t)
=
\mathcal I_Y
+
E_{\theta_0}\{a_t(Y)K(Y)\},
\]
and hence
\begin{equation}
\left.
\frac{d}{dt}\mathcal I(a_t)
\right|_{t=0}
=
E_{\theta_0}\{h(Y)K(Y)\}.
\label{eq:supp-information-derivative}
\end{equation}

For any differentiable nonsingular matrix-valued function \(M(t)\),
\[
\frac{d}{dt}M(t)^{-1}
=
-
M(t)^{-1}
\frac{dM(t)}{dt}
M(t)^{-1}.
\]
Applying this identity to \(M(t)=\mathcal I(a_t)\) and using
\eqref{eq:supp-information-derivative} gives
\begin{align*}
\left.
\frac{d}{dt}\Phi(a_t)
\right|_{t=0}
&=
\left.
\frac{d}{dt}
\operatorname{tr}
\left\{
\widetilde H_R\mathcal I(a_t)^{-1}
\right\}
\right|_{t=0}
\\
&=
-
\operatorname{tr}
\left[
\widetilde H_R
\mathcal I(a)^{-1}
E_{\theta_0}\{h(Y)K(Y)\}
\mathcal I(a)^{-1}
\right].
\end{align*}
By linearity of expectation and trace,
\begin{align*}
\left.
\frac{d}{dt}\Phi(a_t)
\right|_{t=0}
&=
-
E_{\theta_0}
\left[
h(Y)
\operatorname{tr}
\left\{
\widetilde H_R
\mathcal I(a)^{-1}
K(Y)
\mathcal I(a)^{-1}
\right\}
\right]
\\
&=
-
E_{\theta_0}\{h(Y)\psi_a(Y)\},
\end{align*}
which establishes \eqref{eq:gateaux-derivative}.

It remains to obtain the decomposition of \(\psi_a(y)\). From
\eqref{eq:query-information},
\[
K(y)
=
r(y)\widetilde J(y)
+
w(y)q(y)q(y)^\top.
\]
Substituting this expression into \eqref{eq:query-value} and using linearity of
the trace yields
\begin{align*}
\psi_a(y)
&=
r(y)
\operatorname{tr}
\left\{
\widetilde H_R
\mathcal I(a)^{-1}
\widetilde J(y)
\mathcal I(a)^{-1}
\right\}
\\
&\quad+
w(y)
\operatorname{tr}
\left\{
\widetilde H_R
\mathcal I(a)^{-1}
q(y)q(y)^\top
\mathcal I(a)^{-1}
\right\}.
\end{align*}
Using
\[
\operatorname{tr}\{Mqq^\top N\}
=
q^\top NMq
\]
with the symmetry of \(\mathcal I(a)^{-1}\) and
\(\widetilde H_R\), the second term becomes
\[
w(y)
q(y)^\top
\mathcal I(a)^{-1}
\widetilde H_R
\mathcal I(a)^{-1}
q(y).
\]
Therefore
\[
\psi_a(y)
=
\psi_a^L(y)+\psi_a^R(y),
\]
with \(\psi_a^L\) and \(\psi_a^R\) given by
\eqref{eq:label-value} and \eqref{eq:response-value}, respectively.

Finally, since \(w(y)\geq0\), \(\widetilde H_R\succeq0\), and
\(\mathcal I(a)^{-1}\) is symmetric,
\[
\psi_a^R(y)
=
w(y)
\left\|
\widetilde H_R^{1/2}
\mathcal I(a)^{-1}q(y)
\right\|^2
\geq0.
\]
This completes the proof.
\hfill\(\square\)

\subsection{Proof of Proposition~\ref{prop:convexity}}
\label{supp:proof-convexity}

Let \(a_0\) and \(a_1\) be two designs for which the corresponding information
matrices are positive definite, and let
\[
a_t=(1-t)a_0+ta_1,
\qquad 0\leq t\leq1.
\]
By Proposition~\ref{prop:information-decomposition},
\[
\mathcal I(a_t)
=
(1-t)\mathcal I(a_0)+t\mathcal I(a_1).
\]
Set
\[
M_t=\mathcal I(a_t),
\qquad
D=\mathcal I(a_1)-\mathcal I(a_0).
\]
Then
\[
\frac{dM_t}{dt}=D.
\]
Using
\[
\frac{d}{dt}M_t^{-1}
=
-M_t^{-1}DM_t^{-1},
\]
we obtain
\[
\frac{d}{dt}\Phi(a_t)
=
-\operatorname{tr}
\left(
\widetilde H_RM_t^{-1}DM_t^{-1}
\right).
\]
Differentiating once more gives
\begin{align*}
\frac{d^2}{dt^2}\Phi(a_t)
&=
2\operatorname{tr}
\left(
\widetilde H_R
M_t^{-1}D
M_t^{-1}D
M_t^{-1}
\right).
\end{align*}
Let
\[
G=M_t^{-1/2}DM_t^{-1/2},
\qquad
C=M_t^{-1/2}\widetilde H_RM_t^{-1/2}.
\]
Since \(\widetilde H_R\succeq0\), we have \(C\succeq0\), while \(G\) is
symmetric. Cyclic invariance of the trace yields
\[
\frac{d^2}{dt^2}\Phi(a_t)
=
2\operatorname{tr}(CG^2).
\]
Because \(G^2\succeq0\) and \(C\succeq0\),
\[
\operatorname{tr}(CG^2)
=
\operatorname{tr}(C^{1/2}G^2C^{1/2})
\geq0.
\]
Hence
\[
\frac{d^2}{dt^2}\Phi(a_t)\geq0,
\]
which proves convexity along every line segment in the positive-definite
information domain. Therefore \(\Phi\) is convex on any convex subset of that
domain.
\hfill\(\square\)

\subsection{Proof of Theorem~\ref{thm:optimal-acquisition}}
\label{supp:proof-optimal-acquisition}

Let \(a^*\) be optimal and write \(\psi=\psi_{a^*}\).
All probabilities and expectations in this proof are taken under
\(P_{\theta_0}\). For \(\epsilon>0\), choose measurable sets
\(B_0\subseteq\{a^*\leq 1-\epsilon\}\) and
\(B_1\subseteq\{a^*\geq\epsilon\}\) of positive probability, and put
\[
h(y)=\frac{\mathbf{1}_{B_0}(y)}{P_{\theta_0}(B_0)}
     -\frac{\mathbf{1}_{B_1}(y)}{P_{\theta_0}(B_1)}.
\]
Since \(E_{\theta_0}\{h(Y)\}=0\), this exchange preserves the budget.
The slack on \(B_0\) and \(B_1\) ensures that \(a^*+th\) remains in
\([0,1]\) for sufficiently small \(t>0\). Its information matrix
remains positive definite by continuity. The right derivative,
given by the same differentiation as in
Proposition~\ref{prop:risk-query-value}, therefore satisfies
\[
0\leq
\left.\frac{d}{dt}\Phi(a^*+th)\right|_{t=0+}
=
E_{\theta_0}\{\psi(Y)\mid Y\in B_1\}
-
E_{\theta_0}\{\psi(Y)\mid Y\in B_0\}.
\]
It follows that
\[
\operatorname*{ess\,sup}_{\{a^*<1\}}\psi
\leq
\operatorname*{ess\,inf}_{\{a^*>0\}}\psi.
\]
Indeed, if this inequality failed, there would be positive-probability
subsets of \(\{a^*<1\}\) and \(\{a^*>0\}\) on which the scores were
separated in the opposite order. Intersecting these subsets with
\(\{a^*\leq 1-\epsilon\}\) and \(\{a^*\geq\epsilon\}\), respectively,
preserves positive probability for sufficiently small \(\epsilon\),
contradicting the exchange inequality.

Both sets have positive probability because \(0<\rho<1\). Since the
score is nonnegative and integrable, the displayed interval contains
a finite \(\lambda\). For such a \(\lambda\), \(\psi\leq\lambda\)
almost surely on \(\{a^*<1\}\), and \(\psi\geq\lambda\) almost surely
on \(\{a^*>0\}\). These inequalities give the stated threshold
conditions, including equality wherever \(0<a^*<1\).

For sufficiency, let \(a\) be any feasible design and interpret
\(D\Phi(a^*)[a-a^*]\) as the right derivative along the feasible
segment from \(a^*\) to \(a\). By convexity,
\[
\Phi(a)
\geq
\Phi(a^*)+D\Phi(a^*)[a-a^*].
\]
The derivative term is
\[
D\Phi(a^*)[a-a^*]
=
-E_{\theta_0}
\left[
\{a(Y)-a^*(Y)\}\psi_{a^*}(Y)
\right].
\]
Since both designs satisfy the same budget,
\[
E_{\theta_0}\{a(Y)-a^*(Y)\}=0,
\]
and therefore
\[
D\Phi(a^*)[a-a^*]
=
E_{\theta_0}
\left[
\{a(Y)-a^*(Y)\}
\{\lambda-\psi_{a^*}(Y)\}
\right].
\]
If \(a^*(y)=0\), then
\(a(y)-a^*(y)\geq0\) and
\(\lambda-\psi_{a^*}(y)\geq0\).
If \(a^*(y)=1\), then
\(a(y)-a^*(y)\leq0\) and
\(\lambda-\psi_{a^*}(y)\leq0\).
If \(0<a^*(y)<1\), then
\(\lambda-\psi_{a^*}(y)=0\).
Thus the integrand is nonnegative almost surely, so
\[
D\Phi(a^*)[a-a^*]\geq0.
\]
Consequently,
\[
\Phi(a)\geq\Phi(a^*)
\]
for every feasible \(a\), proving global optimality.
\hfill\(\square\)

\subsection{Proof of Proposition~\ref{prop:response-identifiability}}
\label{supp:proof-response-identifiability}

From \eqref{eq:response-nuisance-information},
\[
C_{\eta\eta}(a)
=
E_{\theta_0}
\left[
a(Y)w(Y)q_\eta(Y)q_\eta(Y)^\top
\right].
\]
Therefore, for any \(v\in\mathbb R^q\),
\begin{align*}
v^\top C_{\eta\eta}(a)v
&=
E_{\theta_0}
\left[
a(Y)w(Y)
v^\top q_\eta(Y)q_\eta(Y)^\top v
\right]
\\
&=
E_{\theta_0}
\left[
a(Y)w(Y)
\{v^\top q_\eta(Y)\}^2
\right],
\end{align*}
which proves \eqref{eq:response-identifiability-quadratic}.

Suppose first that there exists \(v\neq0\) such that
\[
v^\top q_\eta(Y)=0
\]
\(\mu_a\)-almost surely. Then
\[
v^\top C_{\eta\eta}(a)v=0,
\]
so \(C_{\eta\eta}(a)\) cannot be positive definite.

Conversely, if \(C_{\eta\eta}(a)\) is not positive definite, its positive
semidefiniteness implies that there exists \(v\neq0\) for which
\[
v^\top C_{\eta\eta}(a)v=0.
\]
Hence
\[
E_{\theta_0}
\left[
a(Y)w(Y)
\{v^\top q_\eta(Y)\}^2
\right]
=0.
\]
The integrand is nonnegative, and therefore
\[
v^\top q_\eta(Y)=0
\]
\(\mu_a\)-almost surely. This establishes the equivalence.

For the model
\[
\operatorname{logit}r(Y;\theta,\eta)
=
\eta_0+\eta_1U(Y;\theta),
\]
we have
\[
q_\eta(Y)=
\begin{pmatrix}
1\\
U(Y;\theta)
\end{pmatrix}.
\]
Let expectation under the normalized measure \(\mu_a\) be denoted by
\(E_{\mu_a}\). Apart from the positive normalizing constant
\(E_{\theta_0}\{a(Y)w(Y)\}\), the nuisance-information matrix is
\[
E_{\mu_a}
\left[
\begin{pmatrix}
1\\ U
\end{pmatrix}
\begin{pmatrix}
1&U
\end{pmatrix}
\right]
=
\begin{pmatrix}
1&E_{\mu_a}(U)\\
E_{\mu_a}(U)&E_{\mu_a}(U^2)
\end{pmatrix}.
\]
Its determinant is
\[
E_{\mu_a}(U^2)-E_{\mu_a}(U)^2
=
\operatorname{Var}_{\mu_a}(U).
\]
Thus, when \(E_{\theta_0}\{a(Y)w(Y)\}>0\),
\[
C_{\eta\eta}(a)\succ0
\quad\Longleftrightarrow\quad
\operatorname{Var}_{\mu_a}\{U(Y;\theta_0)\}>0,
\]
which proves \eqref{eq:uncertainty-identifiability}.
\hfill\(\square\)

\subsection{Proof of Theorem~\ref{thm:adaptive-oracle}}
\label{supp:proof-adaptive-oracle}

\paragraph{Oracle criterion value.}
The approximate minimization condition and the definition of \(\Delta_n\)
give
\[
\Phi(\widehat a_n)
\leq\widehat\Phi_n(\widehat a_n)+\Delta_n
\leq\widehat\Phi_n(a^*)+\varepsilon_n+\Delta_n
\leq\Phi(a^*)+\varepsilon_n+2\Delta_n.
\]
Since \(\widehat a_n\in\mathcal D\) and \(a^*\) minimizes \(\Phi\) there,
\(\Phi(\widehat a_n)\geq\Phi(a^*)\). This proves
\eqref{eq:oracle-criterion-equivalence}. This part does not require
independence of the observations.

\paragraph{Uniform fixed-design second moments.}
For deterministic \(a\in\mathcal D\), put
\[
T_{n,a}=\sqrt n(\widehat\phi_{n,a}-\phi_0),
\qquad
W_{n,a}=\mathcal I(a)^{-1}\frac1{\sqrt n}
\sum_{i=1}^n S_a(O_i;\phi_0).
\]
The iid score identities imply exactly
\[
E_aW_{n,a}=0,
\qquad E_a(W_{n,a}W_{n,a}^\top)=\mathcal I(a)^{-1}.
\]
Writing \(d+q\) for the dimension of \(\phi\), the information lower bound
therefore gives
\[
\sup_{a\in\mathcal D}E_a\|W_{n,a}\|^2
=\sup_{a\in\mathcal D}\operatorname{tr}\{\mathcal I(a)^{-1}\}
\leq(d+q)/\kappa.
\]
By \eqref{eq:uniform-l2-linearization},
\(T_{n,a}=W_{n,a}+\zeta_{n,a}\), with the squared remainder tending to
zero uniformly in expectation. Cauchy--Schwarz applied to the two cross
terms yields
\[
\sup_{a\in\mathcal D}
\left\|E_a(T_{n,a}T_{n,a}^\top)-\mathcal I(a)^{-1}\right\|
\longrightarrow0,
\]
where \(\|\cdot\|\) denotes the operator norm for matrices. With
\(e_{n,a}=\widehat\theta_{n,a}-\theta_0\), taking classification blocks
and using the block-inverse identity gives
\[
\sup_{a\in\mathcal D}
\left\|nE_a(e_{n,a}e_{n,a}^\top)-I_{\mathrm{eff}}(a)^{-1}\right\|
\longrightarrow0.
\]
Consequently,
\[
\sup_{a\in\mathcal D}
\left|nE_a(e_{n,a}^\top H_Re_{n,a})-\Phi(a)\right|
\longrightarrow0.
\]

\paragraph{Uniform control of the risk remainder.}
Define
\[
b(h)=R(\theta_0+h)-R^*-\tfrac12h^\top H_Rh
\]
for parameter perturbations in the domain of the classifier. By
\eqref{eq:risk-expansion}, for every \(\epsilon>0\) there is a radius
\(s>0\) such that \(|b(h)|\leq\epsilon\|h\|^2\) whenever \(\|h\|\leq s\).
Condition \eqref{eq:uniform-estimator-moment} gives a finite constant
\(C\) such that, for all sufficiently large \(n\),
\[
\sup_{a\in\mathcal D}E_a\|\sqrt n e_{n,a}\|^{2+\delta}\leq C.
\]
On the complement of the radius-\(s\) ball, boundedness of classification
risk implies
\(|b(e)|\leq1+\tfrac12\|H_R\|\|e\|^2\). Markov's inequality gives
\[
\begin{aligned}
\sup_a nP_a(\|e_{n,a}\|>s)
&\leq C s^{-(2+\delta)}n^{-\delta/2},\\
\sup_a nE_a\{\|e_{n,a}\|^2
\mathbf1(\|e_{n,a}\|>s)\}
&\leq C s^{-\delta}n^{-\delta/2}.
\end{aligned}
\]
Both bounds tend to zero. Inside the ball,
\[
\sup_a nE_a\{|b(e_{n,a})|\mathbf1(\|e_{n,a}\|\leq s)\}
\leq\epsilon\sup_a E_a\|\sqrt n e_{n,a}\|^2.
\]
The last supremum is bounded, by either the moment condition or the
preceding second-moment calculation. Taking the limit superior and then
letting \(\epsilon\downarrow0\) proves
\[
\sup_{a\in\mathcal D}nE_a|b(e_{n,a})|\longrightarrow0.
\]
Together with the quadratic term, this establishes the uniform expansion
\begin{equation}
\sup_{a\in\mathcal D}
\left|E_a\{R(\widehat\theta_{n,a})-R^*\}
-\frac{\Phi(a)}{2n}\right|=o(n^{-1}).
\label{eq:supp-uniform-fixed-design-risk}
\end{equation}

\paragraph{Conditioning on the pilot.}
For deterministic \(a\), let
\(g_n(a)=E_a\{R(\widehat\theta_{n,a})-R^*\}\).
The selected design is \(\mathcal F_{0,n}\)-measurable, and the estimator
uses the pilot only through this design. The conditional product law
\eqref{eq:conditional-iid-second-stage} therefore gives the exact identity
\[
E\{R(\widehat\theta_{n,\widehat a_n})-R^*
\mid\mathcal F_{0,n}\}=g_n(\widehat a_n).
\]
Set \(r_n=g_n(\widehat a_n)-\Phi(\widehat a_n)/(2n)\).
Equation~\eqref{eq:supp-uniform-fixed-design-risk} implies
\[
E|r_n|
\leq\sup_{a\in\mathcal D}
\left|g_n(a)-\frac{\Phi(a)}{2n}\right|
=o(n^{-1}),
\]
which proves \eqref{eq:conditional-adaptive-risk}. This step is why
uniform fixed-design control is needed instead of an expansion only for
each individual deterministic design.

\paragraph{Averaging over the pilot.}
Taking expectations now gives
\[
E\{R(\widehat\theta_{n,\widehat a_n})\}-R^*
=\frac1{2n}E\{\Phi(\widehat a_n)\}+o(n^{-1}).
\]
On \(\mathcal D\),
\[
0\leq\Phi(a)\leq\kappa^{-1}\operatorname{tr}(\widetilde H_R).
\]
The convergence in probability in
\eqref{eq:oracle-criterion-equivalence}, together with this deterministic
bound, implies \(E\Phi(\widehat a_n)\to\Phi(a^*)\).
This proves \eqref{eq:adaptive-risk-equivalence}. Finally,
\(N_n=n+m_n\) satisfies \(N_n/n\to1\), so replacing \(n\) by \(N_n\)
in the leading denominator changes the expression by \(o(N_n^{-1})\).
This change of denominator concerns sample-size accounting only; the
proof has not treated the pooled two-stage observations as iid.
\hfill\(\square\)


\subsection{Derivation of the binary logistic risk expansion}
\label{supp:logistic-risk-expansion}

For the binary logistic model,
\[
\tau_\theta(y)
=
\frac{\exp(\alpha+\beta y)}
{1+\exp(\alpha+\beta y)},
\qquad
\beta>0,
\]
the Bayes classifier under parameter \(\theta\) changes class at
\[
b(\theta)=-\frac{\alpha}{\beta}.
\]
Let
\[
b_0=b(\theta_0)
=
-\frac{\alpha_0}{\beta_0}.
\]

For \(\theta\) sufficiently close to \(\theta_0\), the classifier based on
\(\theta\) differs from the Bayes classifier only on the interval between
\(b(\theta)\) and \(b_0\). Hence
\begin{equation}
R(\theta)-R^*
=
\left|
\int_{b_0}^{b(\theta)}
\{2\tau_{\theta_0}(y)-1\}f(y)\,dy
\right|.
\label{eq:supp-logistic-risk-integral}
\end{equation}
Since
\[
\tau_{\theta_0}(b_0)=\frac12,
\]
the integrand vanishes at \(b_0\). Moreover,
\[
\frac{d}{dy}\tau_{\theta_0}(y)
=
\beta_0
\tau_{\theta_0}(y)
\{1-\tau_{\theta_0}(y)\},
\]
so
\[
\left.
\frac{d}{dy}
\{2\tau_{\theta_0}(y)-1\}
\right|_{y=b_0}
=
\frac{\beta_0}{2}.
\]
Therefore,
\[
\{2\tau_{\theta_0}(y)-1\}f(y)
=
\frac{\beta_0 f(b_0)}{2}(y-b_0)
+
o(|y-b_0|).
\]
Substitution into \eqref{eq:supp-logistic-risk-integral} gives
\begin{equation}
R(\theta)-R^*
=
\frac{\beta_0 f(b_0)}{4}
\{b(\theta)-b_0\}^2
+
o\left(
\{b(\theta)-b_0\}^2
\right).
\label{eq:supp-boundary-risk}
\end{equation}

Now let
\[
\theta=\theta_0+h,
\qquad
h=
\begin{pmatrix}
h_\alpha\\
h_\beta
\end{pmatrix}.
\]
The gradient of the boundary function is
\[
\nabla_\theta b(\theta)
=
\begin{pmatrix}
-1/\beta\\
\alpha/\beta^2
\end{pmatrix}.
\]
At \(\theta_0\), using
\(\alpha_0=-\beta_0b_0\),
\[
\nabla_\theta b(\theta_0)
=
-\frac{1}{\beta_0}
\begin{pmatrix}
1\\
b_0
\end{pmatrix}.
\]
Thus
\[
b(\theta_0+h)-b_0
=
-\frac{1}{\beta_0}
(h_\alpha+b_0h_\beta)
+
o(\|h\|).
\]
Substituting this expansion into
\eqref{eq:supp-boundary-risk} yields
\[
R(\theta_0+h)-R^*
=
\frac{f(b_0)}{4\beta_0}
(h_\alpha+b_0h_\beta)^2
+
o(\|h\|^2).
\]
Hence
\[
R(\theta_0+h)-R^*
=
\frac12 h^\top H_Rh+o(\|h\|^2),
\]
with
\[
H_R
=
\frac{f(b_0)}{2\beta_0}
\begin{pmatrix}
1\\
b_0
\end{pmatrix}
\begin{pmatrix}
1&b_0
\end{pmatrix}.
\]
This proves \eqref{eq:logistic-risk-expansion} and
\eqref{eq:logistic-risk-curvature}.
\hfill\(\square\)


%

\section{Response-term ablation}
\label{supp:response-term-ablation}

The ablation uses the full query--response information matrix and
the same joint likelihood as response-aware acquisition, but omits
\(\psi_a^R\) from candidate scores.
All comparisons use 500 paired replications and total query budgets
including the common pilot.

For each replication, the paired difference is the excess risk of
the full criterion minus that of the ablation.
Intervals are pointwise 95\% Monte Carlo \(t\) intervals.
Two-sided paired \(t\)-test \(p\)-values are adjusted by Holm's method
over the nine regime--budget comparisons separately within the oracle
and adaptive studies.
The oracle null comparisons are exactly zero; their intervals are
degenerate and their \(p\)-values are recorded as one by convention.

\begin{table*}[htbp]
\centering
\caption{Paired excess-risk differences for response-aware acquisition
versus the label-only, full-information ablation.
Differences and interval endpoints are multiplied by \(10^3\).
Negative values favor response-aware acquisition.}
\label{tab:supp-response-term-ablation}

\small
\setlength{\tabcolsep}{4pt}
\renewcommand{\arraystretch}{1.05}

\resizebox{\textwidth}{!}{%
\begin{tabular}{lllrrr}
\toprule
Study & Response & \(B\)
& Difference & 95\% interval & \(p_{\mathrm{Holm}}\) \\
\midrule

Oracle & Noninformative & 300
& 0.0000 & [0.0000, 0.0000] & 1.0000 \\
& & 600
& 0.0000 & [0.0000, 0.0000] & 1.0000 \\
& & 900
& 0.0000 & [0.0000, 0.0000] & 1.0000 \\
\addlinespace

& Moderate & 300
& -0.0874 & [-0.3741, 0.1994] & 1.0000 \\
& & 600
& -0.0960 & [-0.2255, 0.0335] & 1.0000 \\
& & 900
& -0.1107 & [-0.1862, -0.0352] & 0.0373 \\
\addlinespace

& Strong & 300
& -0.0372 & [-0.1182, 0.0437] & 1.0000 \\
& & 600
& 0.0013 & [-0.0162, 0.0188] & 1.0000 \\
& & 900
& -0.0071 & [-0.0198, 0.0056] & 1.0000 \\
\midrule

Adaptive & Noninformative & 300
& 0.2192 & [-0.1308, 0.5691] & 1.0000 \\
& & 600
& 0.0707 & [-0.0577, 0.1990] & 1.0000 \\
& & 900
& 0.0354 & [-0.0181, 0.0889] & 1.0000 \\
\addlinespace

& Moderate & 300
& -0.2205 & [-0.4623, 0.0214] & 0.5912 \\
& & 600
& -0.0832 & [-0.2036, 0.0372] & 1.0000 \\
& & 900
& -0.1077 & [-0.1726, -0.0428] & 0.0107 \\
\addlinespace

& Strong & 300
& -0.0305 & [-0.1247, 0.0638] & 1.0000 \\
& & 600
& -0.0259 & [-0.0602, 0.0083] & 0.9648 \\
& & 900
& -0.0071 & [-0.0247, 0.0104] & 1.0000 \\
\bottomrule
\end{tabular}%
}
\end{table*}

\begin{figure}[htbp]
\centering
\includegraphics[width=\textwidth]{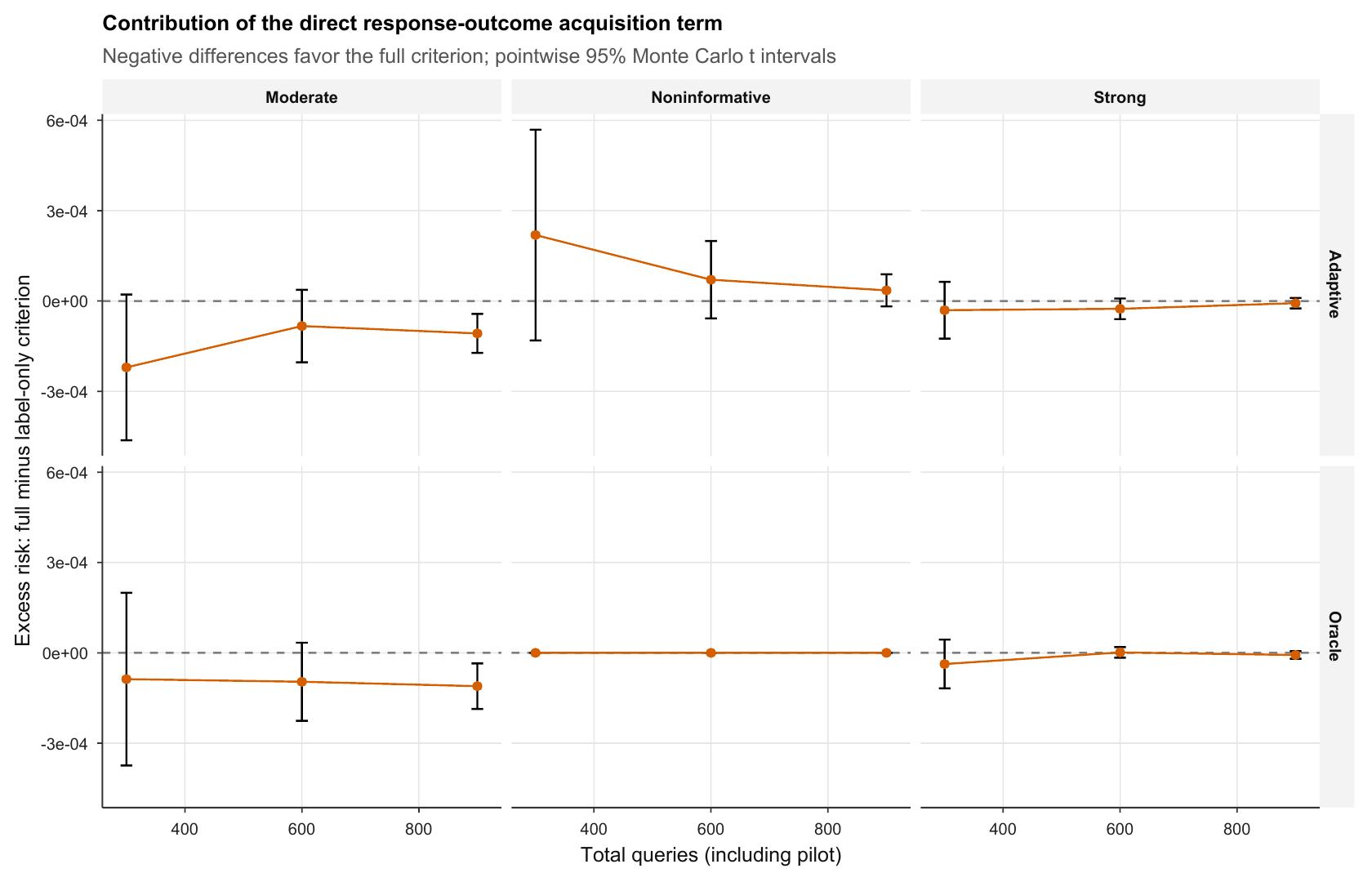}
\caption{ Excess-risk differences between the full response-aware
criterion and the label-only, full-information ablation.
Negative differences favor the full criterion.
Bars are pointwise 95\% Monte Carlo \(t\) intervals over
500 paired replications. Query budgets include the pilot.}
\label{fig:supp-response-term-ablation}
\end{figure}

Figure~\ref{fig:supp-response-term-ablation} displays the paired
differences and pointwise intervals.
The intervals are not simultaneous confidence intervals.

\end{document}